%% file: main.tex
\documentclass[10pt, a4paper]{article}
\usepackage{times}  
\usepackage{helvet} 
\usepackage{courier} 
\usepackage[hyphens]{url} 
\usepackage{graphicx} 
\usepackage{caption} 
\usepackage{siunitx}
\usepackage{subfig}
\usepackage{microtype}
\usepackage{lmodern}
\usepackage{booktabs}       
\usepackage{hyperref}
\usepackage{comment}
\usepackage{amsthm}
\usepackage{amsfonts}
\usepackage{mathtools}
\usepackage{algorithm}
\usepackage{algorithmic}
\usepackage[table]{xcolor}
\usepackage[utf8]{inputenc} 
\usepackage[T1]{fontenc}    
\usepackage{nicefrac}       
\usepackage{wrapfig}
\usepackage{subfig}
\usepackage{multirow} 

\newtheorem{theorem}{Theorem}
\newtheorem{lemma}{Lemma}
\newtheorem{definition}{Definition}

\PassOptionsToPackage{square,numbers}{natbib}
\usepackage{natbib}  

\allowdisplaybreaks

\newcommand{\round}[2]{\num[round-mode=places,round-precision=#1]{#2}}
\newcommand{\lvl}{~~~}

\definecolor{grayl}{gray}{0.3}
\definecolor{graym}{gray}{0.5}
\definecolor{grays}{gray}{0.7}

\usepackage{amsmath,amsfonts,bm}
\usepackage{amssymb, wasysym}

\newcommand{\transpose}{^\mathsf{T}}

\def\vone{{\bm{1}}}
\newcommand{\textF}{\text{F}}
\newcommand{\texts}{\text{s}}
\newcommand{\textt}{\text{t}}
\newcommand{\textd}{\text{d}}
\newcommand{\euc}{\text{E}}
\newcommand{\hyp}{\text{H}}

\DeclareMathOperator{\Sym}{Sym}
\DeclareMathOperator{\diag}{diag}

\DeclareMathOperator{\gyr}{gyr}

\def\eqref#1{equation~\ref{#1}}

\def\1{\bm{1}}

\def\eps{{\epsilon}}

\newcommand{\Exp}{\textnormal{Exp}}
\newcommand{\Log}{\textnormal{Log}}

\newcommand{\recloss}{l}

\def\rva{{\mathbf{a}}}
\def\rvb{{\mathbf{b}}}

\def\rvg{{\mathbf{g}}}
\def\rvh{{\mathbf{h}}}
\def\rvu{{\mathbf{i}}}

\def\rvu{{\mathbf{u}}}
\def\rvv{{\mathbf{v}}}
\def\rvw{{\mathbf{w}}}
\def\rvx{{\mathbf{x}}}
\def\rvy{{\mathbf{y}}}
\def\rvz{{\mathbf{z}}}

\def\rmA{{\mathbf{A}}}
\def\rmB{{\mathbf{B}}}
\def\rmC{{\mathbf{C}}}

\def\rmF{{\mathbf{F}}}
\def\rmG{{\mathbf{G}}}

\def\rmI{{\mathbf{I}}}

\def\rmK{{\mathbf{K}}}
\def\rmL{{\mathbf{L}}}
\def\rmM{{\mathbf{M}}}

\def\rmP{{\mathbf{P}}}
\def\rmQ{{\mathbf{Q}}}

\def\rmT{{\mathbf{T}}}

\def\rmW{{\mathbf{W}}}
\def\rmX{{\mathbf{X}}}

\def\vzero{{\bm{0}}}
\def\vone{{\bm{1}}}

\DeclareMathAlphabet{\mathsfit}{\encodingdefault}{\sfdefault}{m}{sl}
\SetMathAlphabet{\mathsfit}{bold}{\encodingdefault}{\sfdefault}{bx}{n}

\def\gM{{\mathcal{M}}}
\def\gN{{\mathcal{N}}}

\def\gP{{\mathcal{P}}}

\def\gT{{\mathcal{T}}}

\def\gX{{\mathcal{X}}}
\def\gY{{\mathcal{Y}}}
\def\gZ{{\mathcal{Z}}}

\def\sB{{\mathbb{B}}}

\def\sR{{\mathbb{R}}}

\newcommand{\E}{\mathbb{E}}

\DeclareMathOperator*{\argmin}{arg\,min}

\DeclareMathOperator{\Tr}{Tr}

\title{Aligning Hyperbolic Representations: an Optimal Transport-based approach}

\author{%
	Andrés Hoyos-Idrobo\\
	Rakuten Institute of Technology, Paris}

\begin{document}

\maketitle

\begin{abstract}
Hyperbolic-spaces are better suited to represent data with underlying hierarchical relationships, e.g., tree-like data. 
However, it is often necessary to incorporate, through alignment, different but related representations meaningfully. 
This aligning is an important class of machine learning problems, with applications as ontology matching and cross-lingual alignment. Optimal transport (OT)-based approaches are a natural choice to tackle the alignment problem as they aim to find a transformation of the source dataset to match a target dataset, subject to some distribution constraints. 
This work proposes a novel approach based on OT of embeddings on the Poincaré model of hyperbolic spaces. 
Our method relies on the gyrobarycenter mapping on M\"obius gyrovector spaces. 
As a result of this formalism, we derive extensions to some existing Euclidean methods of OT-based domain adaptation to their hyperbolic counterparts. 
Empirically, we show that both Euclidean and hyperbolic methods have similar performances in the context of retrieval. 
\end{abstract}

\section{Introduction}
Hyperbolic embeddings are state-of-the-art models to learn representations of data with an underlying hierarchical structure~\citep{de2018representation}.  
The hyperbolic space serves as a geometric prior to hierarchical structures, tree graphs, heavy-tailed distributions, e.g., scale-free, power-law~\cite{nickel2017poincare}.
A relevant tool to implement hyperbolic space algorithms is the M\"obius gyrovector spaces or Gyrovector spaces~\cite{ungar2008gyrovector}.
Gyrovector spaces are an algebraic formalism, which leads to vector-like operations, i.e., gyrovector, in the Poincaré model of the hyperbolic space. 
Thanks to this formalism, we can quickly build estimators that are well-suited to perform end-to-end optimization~\cite{becigneul2018riemannian}. 
Gyrovector spaces are essential to design the hyperbolic version of several machine learning algorithms, like
Hyperbolic Neural Networks (HNN)~\citep{ganea2018hyperbolic}, 
Hyperbolic Graph NN~\citep{liu2019hyperbolic}, 
Hyperbolic Graph Convolutional NN~\citep{chami2019hyperbolic}, 
learning latent feature representations~\citep{mathieu2019continuous, ovinnikov2019poincar},  
word embeddings~\citep{tifrea2018poincar, ganea2018hyperbolicCones}, 
and image embeddings~\citep{khrulkov2019hyperbolic}.

Modern machine learning algorithms rely on the availability to accumulate large volumes of data, often coming from various sources, e.g., acquisition devices or languages. 
However, these massive amounts of heterogeneous data can entangle downstream learning tasks since the data may follow different distributions. 
Alignment aims at building connections between two or more disparate data sets by aligning their underlying manifolds. 
Thus, it leverages information across datasets. 
This aligning is an important class of machine learning problems, with applications as ontology matching and cross-language information retrieval~\cite{alvarez2019unsupervised}.

Nowadays, optimal transport (OT)-based similarity measures are well-suited in matching  datasets  tasks as they preserve the topology of the data~\cite{peyre2019computational}.
They estimate a transformation of the source domain sample that minimizes their average displacement to the target domain sample. 
These methods assume the existence of an unknown function $T$ in a pre-specified class $\gT$, 
which characterizes the global correspondence between spaces source and target.
In practice, we are interested in learning this transformation.
Typically, the selection of $\gT$ has to be informed by the application domain. 
There are several studies in OT-based aligning in the context of domain adaptation in Euclidean 
spaces~\cite{courty2016optimal, perrot2016mapping,seguy2018large}.
Recently, \cite{alvarez2019unsupervised}  proposed to use OT-based similarity 
measures and a HNN as an estimator to align hyperbolic embeddings.
Nevertheless, they mention the lack of stability of this approach due to the network's initialization procedure, which is something we tackle in this paper. 

\paragraph{Our contributions.}
In this work, we propose various novel approaches to align Poincaré embeddings based on OT. 
In particular,  we rely on the gyrobarycenter mapping (GM) on Möbius gyrovector spaces to derive extensions to some existing Euclidean methods of OT-based domain adaptation to their hyperbolic counterparts. 
To our knowledge, we provide a first theoretical analysis of the consistency of the hyperbolic linear layer for the regularized hyperbolic regression problem. 
We establish a link within the hyperbolic linear layer and wrapped Gaussian distributions. 
We explore GM as a strategy to increase the stability of end-to-end minimization of the current hyperbolic OT-based approach.

\paragraph{Notation.} 
We denote vectors as bold lower-case, e.g, $\mathbf{a}$.
We write matrices using bold capital letters, e.g., $\mathbf{A}$.
Let $\|\cdot\|$ be the $\ell_2$ norm of a vector. 
$\mathbf{A}^\top$ is the transpose of $\mathbf{A}$.
Letters in calligraphic, e.g. $\gP$ denotes sets.
We denote the $(m - 1)$-dimensional probability simplex by 
$\Delta^m$. 
Let $n$ be the number of data points and $[n]$ denotes $ \{1, \ldots, n\}$. 
$\vone_{n}$ is a $n$-dimensional vector of ones (similarly for $\vzero$).
Let $\rmI$ be the identity matrix, and $\langle \rmA,\, \rmB\rangle_{\textF} = \Tr(\rmA\transpose \rmB)$.

\section{Background}
\subsection{Optimal transport\label{subsec:optimal_transport}}
\paragraph{The Monge problem.}
For two random variables $\rvx^\texts \sim \alpha$  and $\rvx^\textt \sim \beta$ with values in $\gX^\texts$ and $\gX^\textt$, respectively. 
The Monge problem seeks a transport map $T: \gX^\texts \rightarrow \gX^\textt$ that 
assigns each point $\rvx^\texts$  to a single point $\rvx^\textt$,  which pushes the mass of $\alpha$ towards that of $\beta$ (i.e., $T_\sharp  \alpha = \beta$), 
while minimizing the transportation cost $c$, e.g., the squared Euclidean distance, as follows:
\begin{equation}
\label{eq:monge_problem}
\argmin_{T} \left\{\E\left[c(\rvx^\texts, T\left(\rvx^\texts\right))\right]|\, T_\sharp \alpha = \beta \right\}, \quad c: \gX^\texts \times \gX^\textt \rightarrow \mathbb{R}^{+}.
\end{equation}

We take $\alpha$ and $\beta$ as empirical measures:
$\alpha = \sum_{i=1}^{n^\texts} \rva_i \delta_{\rvx_i^\texts}$ and
$\beta = \sum_{j=1}^{n^\textt} \rvb_j \delta_{\rvx_j^\textt}$, 
where $\delta_\rvx$ is the Dirac at position $\rvx$, 
$\rva \in \Delta^{n^\texts}$ and  $\rvb \in \Delta^{n^\textt}$ are the weights.
When $\alpha$ is a discrete measure, a map $T$ satisfying the constraint may not exist, e.g., when the target measure has more points than the source measure~\citep{peyre2019computational}.

\paragraph{Relaxation and regularization.}
The Monge problem is not always relevant to studying discrete measures (i.e., histograms) like the ones presented in this paper.
The idea of the Kantorovich relaxation is to use ``soft'' assignments defined in terms of probabilistic couplings 
$\Pi(\rva, \rvb) = \left\{\rmM \in \sR_+^{n^\texts \times n^\textt} |\, \rmM\,\vone = \rva, \, \rmM^\top\, \vone = \rvb \right\}$
that solves 
\begin{equation}
\label{eq:kantorovich_problem}
\text{OT}(\rva, \rvb) \coloneqq \argmin_{\rmM \in \Pi(\rva, \rvb) } \left\langle \rmM,\, \rmC\right\rangle_{\textF},
\end{equation}
where $\rmC \in \sR^{n^\texts\times n^\textt}$ is the cost matrix related to the cost function $c$. 
The discrete optimal transport problem is a linear program that can be solved using specialized algorithms such as the network simplex or interior-point-methods. However, the current implementation of these algorithms has cubic complexity in the size of the support of  $\alpha$ and $\beta$~\citep{pele2009fast}.
\cite{cuturi2013sinkhorn} proposed to add an entropic regularization to speed-up computations, namely 
\begin{equation}
\label{eq:sinkhorn}
W_\epsilon(\rva, \rvb) \coloneqq \min_{\rmM \in \Pi(\rva, \rvb)} \langle \rmM,\, \rmC\rangle_{\textF} - \epsilon\, H(\rmM),
\end{equation}
where  $\epsilon \geq 0$ is the regularization parameter, and $H(\rmM)$ denotes 
the discrete entropy of a coupling matrix~\citep{peyre2019computational}, 
$H(\rmM) = - \sum_{i, j} \rmM_{i, j} \left(\log\left( \rmM_{i, j} \right) -1 \right)$, 
with the convention: $H(a) = -\infty$ if $a$ is $0$ or negative.
The Sinkhorn algorithm~\citep{cuturi2013sinkhorn} solves Eq.~\ref{eq:sinkhorn} and  
can be differentiated using automatic differentiation.
However, Eq.~\ref{eq:sinkhorn} is not a divergence, i.e., $W_\epsilon(\alpha,\, \alpha) \neq 0$.
\cite{genevay2018learning} proposes a divergence that overcomes the issues of the entropic regularization, the Sinkhorn divergence:
$SD_\epsilon(\alpha, \beta) \coloneqq W_\epsilon(\alpha, \beta) - \frac{1}{2}\left(W_\epsilon(\alpha, \alpha) + W_\epsilon(\beta, \beta) \right)$.

In practice, we encode supervision by enforcing $\rmC_{i, j} = 0$ whenever $\rvx_i^\texts$ matches $\rvx_j^\texts$ and $\infty$ otherwise~\cite{courty2016optimal}.

\paragraph{Barycenter mapping.}
Once we obtain a transport coupling $\rmM$ using either Eq.~\ref{eq:kantorovich_problem} or \ref{eq:sinkhorn},  
we can express the transported samples from the source as barycenters of the target samples~\citep{ferradans2014regularized, courty2014domain, perrot2016mapping}. 
This barycentric projection corresponds to solving 
$\hat{\rvx}_i^\texts = \argmin_{\rvx \in \gX^\textt} \sum_{j=1}^{n^\textt} \rmM_{ij} \, c(\rvx, \, \rvx_j^\textt)$, 
which has a closed-form solution 
if $c$ is the squared $\ell_2$ distance, and it corresponds to a weighted average:
\begin{equation}
\label{eq:euclidean_barycenter_map}
\hat{\rmX}^\texts = B_{\rmM}^{\euc} (\rmX^\texts) = \diag(\rmM\, \vone_{n^\textt})^{-1}\, \rmM\, \rmX^\textt.
\end{equation}
In particular, assuming uniform sampling, it reduces to $\hat{\rmX}^\texts =  n^\texts\, \rmM\, \rmX^\textt$.
Thus, we map the source sample into the convex hull of the target examples. 
We use the barycenter mapping as a reference to learn an approximate transport function~\citep{perrot2016mapping,seguy2018large}.

\subsection{Learning in Hyperbolic Spaces~\label{sec:hyperbolic_spaces}}
\paragraph{Hyperbolic spaces.}
The hyperbolic space is a Riemannian manifold of constant negative curvature.
A Riemannian manifold is a smooth manifold $\gM$ of dimension $\textd$ 
that can be locally approximated by an Euclidean space $\sR^\textd$.
At each point $\rvx \in \gM$ one can define  a $\textd$-dimensional vector space, the tangent space $T_{\rvx}\gM$.
We characterize the structure of this manifold by a Riemannian metric, which is a collection of
scalar products $\rho = \left\{\rho(\cdot,\, \cdot)_{\rvx}\right\}_{\rvx \in  \gM}$,
where $\rho(\cdot,\, \cdot)_{\rvx}: T_{\rvx}\gM \times T_{\rvx}\gM  \rightarrow \sR$. 
%
The Riemannian manifold is a pair $(\gM, \rho)$~\citep{sommer2020introduction,pennec2011riemannian}.

The tangent space linearizes the manifold at a point $\rvx \in \gM$, making it suitable for practical applications 
as it leverages the implementation of algorithms in the Euclidean space. 
We use the Riemannian exponential and logarithmic maps to project samples
onto the manifold and back to tangent space, respectively.
The Riemannian exponential map, when well-defined, $\Exp_{\rvx}: T_{\rvx}\gM \rightarrow \gM$ 
realizes a local diffeomorphism from a sufficiently small neighborhood $\vzero$ in $T_{\rvx}\gM$ into
a neighborhood of the point $\rvx \in \gM$. 
The logarithmic map is the inverse exponential map, $\Log_{\rvx}: \gM  \rightarrow T_{\rvx}\gM$.

There are five isometric models of the hyperbolic manifold~\citep{cannon1997hyperbolic}.
Our focus is on the Poincaré ball model because of the M\"obius gyrovector formalism (see below).
The Poincaré ball 
$\sB_s^\textd = \left\{ \rvx \in \sR^\textd:\,  \|\rvx\|^2 < s, \, s\geq 0 \right\}$ has the Riemannian metric $\rho_\rvx^{\sB_s} = (\lambda_{\rvx}^s)^2 \rho^\text{E}$, where  
$\lambda_{\rvx}^s \coloneqq 2 / \left( 1 -  \|\rvx\|^2/s\right)$,
and $\rho^\text{E} = \rmI_\textd$ is the Euclidean metric tensor.
Hence, the conformal factor $\lambda_{\rvx}^s$~\citep{pennec2011riemannian} connects the Euclidean space and the Poincaré ball.

\paragraph{Gyrovector formalism.}
%
%
Gyrovector spaces analogize Euclidean vector spaces and provide a non-associative
algebraic formalism for the hyperbolic geometry of the Poincar\'e model~\cite{ungar2008gyrovector,ganea2018hyperbolic}.  
Let $(\sB_s^\textd, \oplus_s, \otimes_s)$ be a M\"{o}bius gyrovector space with 
M\"{o}bius addition $\oplus_s$ and M\"{o}bius product $\otimes_s$.
These operations and their properties are described in closed-forms in Section 4 in supp.~mat. 
A remark, the M\"{o}bius addition $\oplus_s$ is often non-commutative, $-(\rvu \oplus_s \rvv) \oplus_s \rvu \neq \rvv$.
However, we can subtract using this operation from the left, $-\rvu \oplus_s (\rvu \oplus_s \rvv) = \rvv$. 
This is known as the left-cancellation law~\cite{ungar2008gyrovector}.

The Poincaré distance $d_s: \sB_s^\textd\times \sB_s^\textd \rightarrow \sR_+$~\citep{ungar2014analytic} is:
\begin{equation}
	d_s(\rvx, \rvy) = 2s\, \tanh^{-1}\left(\frac{\|-\rvx \oplus_s \rvy\|}{s}\right).
\end{equation}

We use wrap functions to compute functions on the manifold~\citep{ganea2018hyperbolic}.  
Let $f: \sR^\textd \rightarrow \sR^\text{p}$ then the M\"obius version of $f$ that maps from $\sB^\textd$ to $\sB^\text{m}$ is defined by:
\begin{equation}
\label{eq:mobius_apply_function}
f^{\otimes_s} (\rvx) = \Exp_{\vzero}\left(f(\Log_{\vzero}(\rvx))\right).
\end{equation}
We obtain the M\"obius matrix-vector multiplication 
by using a linear map $\rmQ: \sR^\textd \rightarrow \sR^\text{p}$ 
in Eq.~\ref{eq:mobius_apply_function}, 
$\rmQ^{\otimes_s} \rvx$.

\paragraph{Hyperbolic neural networks (HNN).}
As their Euclidean counterpart, hyperbolic neural networks~\citep{ganea2018hyperbolic} rely on a sequence of matrix-vector 
multiplications followed by an element-wise non-linearity.
For $i> 0$ hidden layers, 
\begin{equation}
\label{eq:hyperbolic_neural_network}
\rvh^{(i)} = \sigma_{\sB}\Biggl( \underbrace{\rvb^{(i)} \oplus_s \left(\left(\rmW^{(i)}\right)^{\otimes_s} \rvh^{(i - 1)} \right)}_{\text{Hyperbolic linear layer}}\Biggr),
\end{equation}
where $\rmW^{(i)} \in \sR^{\textd^{(i)} \times \textd^{(i-1)}}$ is the weight matrix, $\rvb^{(i)} \in \sB_s^{\textd^{(i)}}$ the bias term, and $\sigma_\sB(\cdot)$ 
is a M\"obius function in the form of Eq.~\ref{eq:mobius_apply_function} with a standard non-linearity, e.g., ReLU.
The intermediate layers of the network do not need to have the same dimension. 
Here, we use the left M\"obius addition instead of the right one~\cite{ganea2018hyperbolic}.
As for our analysis, we rely heavily on the left cancellation-rule
(See Section 5 in supp.~mat.).

\paragraph{Gyrobarycenter mapping (GM).}
The gyrobarycentric coordinates are the hyperbolic version of the Euclidean barycentric coordinates. 
The hyperbolic graph neural networks~\cite{gulcehre2018hyperbolic, khrulkov2019hyperbolic, bachmann2019constant} use GM in the aggregation step.
We propose a matrix version of the gyrobarycentric coordinates in M\"obius Gyrovector Spaces~\citep{albert2010barycentric},
\begin{equation}
\label{eq:gyrobarycenter}
\begin{aligned}
B_{\rmM}^\hyp (\rmX^\texts) = \frac{1}{2} \otimes_s  \diag\left(\rmM\, \rvg \right)^{-1}\,\rmM\, \rmG \, \rmX^\textt,
\end{aligned}
\end{equation}
where $\rvg = (\gamma_{\rmX^\textt}^s)^2 - \frac{1}{2}$, $\rmG = \diag((\gamma_{\rmX^\textt}^s)^2)$, 
and $\gamma_{\rmX^\textt}^s$ denotes the Lorentz gamma factor applied sample-wise.
We observe that $\lim_{\texts \rightarrow \infty} B_{\rmM}^{\hyp} (\rmX^\texts) = \frac{4}{7}\, B_{\rmM}^{\euc} (\rmX^\texts)$, 
i.e., we recover Euclidean barycenter in the limit\footnote{
	The factor $\frac{7}{4}$ comes form the distortion induced by the conformal factor $\gamma_{\rmX}^s$. 
	See Section 5 in supp.~mat.}.

GM is a geodesically-convex combination of target samples.
Empirically, the gyromidpoint (i.e., gyrobarycenter with uniform weights) 
is close to the Fréchet/Karcher mean~\cite{chamberlain2019scalable}.
The Fréchet mean is the point with minimum weighted mean squared-Riemannian 
distance to all 
the set~\cite{frechet1948elements}.
Contrary to the gyrobarycenter, it has no closed-solution. 
We rely on iterative algorithms to compute the Fr\'echet mean~\cite{Pennec2020}, 
which can be prohibitively expensive, especially when one requires gradients to flow through the solution.

\begin{figure}[h]
	\centering
	\includegraphics[width=.35\linewidth, trim={0 0mm 0mm 0mm}, clip]{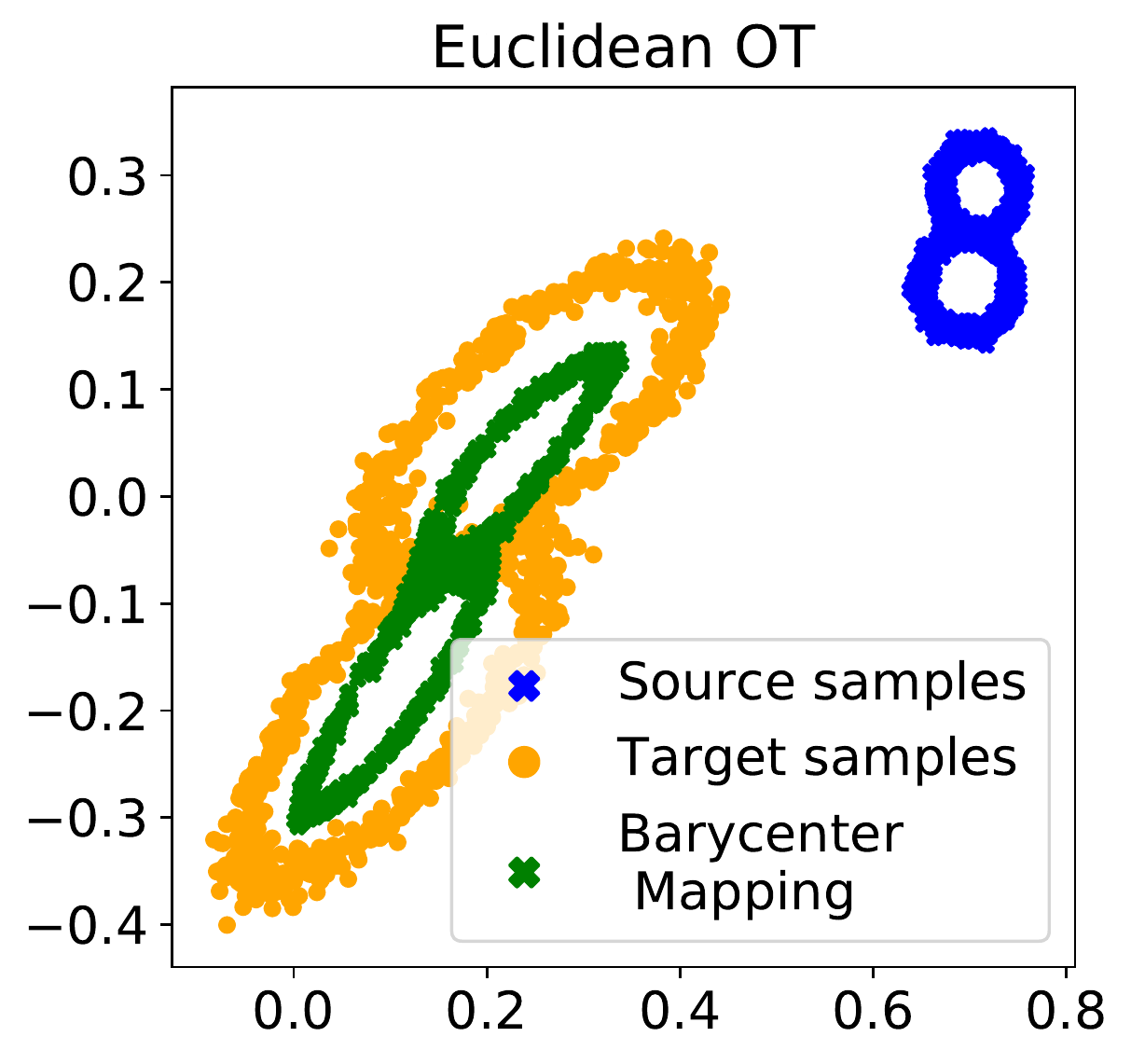}
	\includegraphics[width=.35\linewidth, trim={0 0mm 0mm 0mm}, clip]{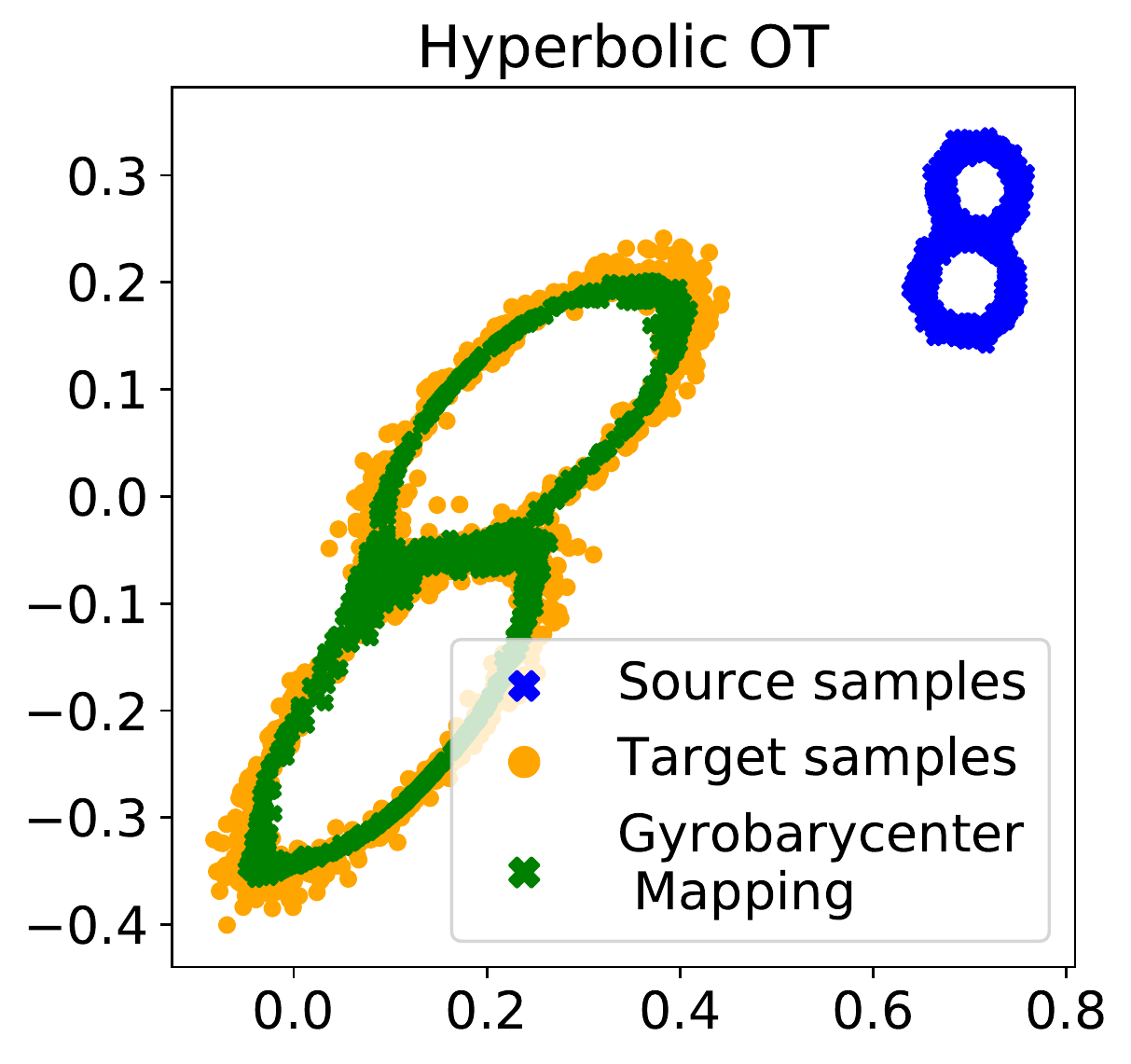}
	\caption{\small \textbf{Illustration:}
		We obtain the probabilistic coupling using Euclidean OT-problem \emph{(left)} and 
		Hyperbolic OT-problem \emph{(right)}.
		We set $\epsilon=0.01$ for both methods.
	}
	\label{fig:toy_hyperbolic_ot}
\end{figure}

\paragraph{Hyperbolic OT.}
We introduced OT problems in Section~\ref{subsec:optimal_transport}, 
which defines the minimum cost assignment between two Euclidean spaces. 
%
Then, the question is whether or not OT extends to hyperbolic spaces.
The answer is partially positive~\citep{alvarez2019unsupervised, mccann2011five}. 
In hyperbolic spaces, there is no guarantee for the smoothness 
of the transport map for the usual cost $d_s\left(\rvx, \rvy\right)^2$ for $\rvx, \rvy \in \sB_s^\textd$.
However, we use compositions of a function $l$ with the Riemannian distance, 
e.g., $-\cosh \circ\, d$~\citep{li2009smooth,lee2009new}. 
We provide more details in Section 3 of supp.~mat.
Fig.~\ref{fig:toy_hyperbolic_ot} shows an illustration of the Euclidean and hyperbolic OT.

\section{Related work~\label{sec:related-work}}
\paragraph{Mapping estimation (ME).}
These methods avoid direct minimization of OT-based similarity measures, 
like Eq.~\ref{eq:sinkhorn}  or the Sinkhorn divergence. 
Instead, they alternate between approximating the barycenter projection and the transport plan.
%
ME uses different and often problem-dependent estimators for the approximation.
Some estimators in the literature are Nearest neighbors (OT domain adaptation, OT-DA)\cite{courty2016optimal}, 
Kernel methods~\cite{perrot2016mapping}, and Neural Networks~\cite{seguy2018large}.
A particular case is the linear mapping, which has closed-solution~\cite{peyre2019computational, flamary2019concentration}. 
As far as we know, there is no adaptation/implementation of these approaches in hyperbolic spaces, which is part of our contribution.

\paragraph{Direct minimization of OT-loss (OT-Direct).}
These methods minimize the OT-based similarity measure between target samples and transformed source samples. 
This idea is present in the Euclidean case in several OT-based algorithms for domain adaptation~\citep{courty2017joint,damodaran2018deepjdot}.
More recently, \cite{alvarez2019unsupervised} proposes an unsupervised approach to 
match points in two different Poincar\'e spaces. 
\cite{alvarez2019unsupervised} optimizes the OT-loss and the transport map simultaneously, using as a loss either 
$W_\epsilon(\alpha, \, T_\sharp(\beta))$ or $SD_\epsilon(\alpha, \, T_\sharp(\beta))$ with a suitable hyperbolic cost, where $T$ is a HNN. 
%
However, this approach is sensitive to initialization due to both the objective function and HNN's nature~\cite{ganea2018hyperbolicCones}.
They propose three pre-training initialization to increase robustness to initialization: 
\begin{itemize}
\item \textbf{Permutation.} The map approximately matches a random permutation of target samples:
\begin{equation}
\label{eq:permutation}
	T \leftarrow \argmin_{T} \frac{1}{n^\texts}\sum_{i=1}^{n^\texts} d_s(\rvx_{\sigma(i)}^\textt, \, T(\rvx_i^\texts)),
\end{equation} 
for some permutation $\sigma(i)$.
\item \textbf{Identity.} The map approximately preserves the geometry of the source space:
\begin{equation}
\label{eq:identity}
T \leftarrow \argmin_{T} \frac{1}{n^\texts} \sum_{i=1}^{n^\texts} d_s(\rvx_i^\texts, \, T(\rvx_i^\texts)).
\end{equation} 
\item \textbf{Procrustes.} The map is approximately a rotation of the source space:
\begin{equation}
\label{eq:procustes}
T \leftarrow \argmin_{T} \frac{1}{n^\texts}\sum_{i=1}^{n^\texts} d_s(\rmP\,\rvx_i^\texts, \, T(\rvx_i^\texts)),
\end{equation} 
 where $\rmP = \argmin_{\rmP\transpose \rmP = \rmI} \|\rmX^\textt - \rmP\,\rmX^\texts\|_{\textF}^2$.
\end{itemize}

Here, we introduce the GM as initialization strategy
to reduce the lack of stability due to initialization:
\begin{equation}
\label{eq:gm_initialization}
T \leftarrow \argmin_{T} \recloss^\hyp(T,\, B_{\rmM}^\hyp) \coloneqq \frac{1}{n^\texts}\sum_{i=1}^{n^\texts} 
d_s\left(B_\rmM^\hyp(\rvx_i^\texts), \,T(\rvx_i^\texts)\right),
\end{equation} 
where $\recloss^\hyp(T,\, B_{\rmM}^\hyp)$ is the fitting data term, 
which assess the quality of the map to reconstruct a given GM.
Our motivation follows Procrustes initialization. 
Instead of finding a rotation, we use the GM which, represents a possibly highly nonlinear approximation encoded in the probabilistic coupling $\rmM$. 
The GM initialization requires solving Eq.~\ref{eq:kantorovich_problem} or~\ref{eq:sinkhorn}, making it significantly more computationally expensive than the other three pre-training approaches.

\section{Aligning Hyperbolic Representations}

\subsection{Hyperbolic OT-DA~\label{sec:hyp_OT_DA}}
We also extend the Euclidean OT-DA algorithm~\cite{courty2016optimal,ferradans2014regularized} to its hyperbolic version.
The most straightforward OT-DA algorithm uses a nearest-neighbor interpolation approach. 
While training, this algorithm memorizes the barycenter mapping of a given set of source samples. 
If a new sample does not belong to the training set, 
the algorithm uses the nearest-neighbor interpolation of the barycenter mapping.
In our extension, we replace the barycenter by the GM and rely on a M\"obius-based nearest-neighbor interpolation. 
Hence, 
$T(\rvx) = B_\rmM^\hyp\left(\rmX_{i(\rvx)}^\texts\right) \oplus _s\left(-\rmX_{i(\rvx)}^\texts \oplus_s \rvx\right)$
where $i(\rvx) = \argmin_{i\in [n^{\texts}]} d_s(\rvx,\, \rmX_i^\texts)$.

\subsection{Hyperbolic Mapping Estimation~\label{sec:hyp_ME}}
Our hyperbolic mapping estimation (Hyp-ME) definition is inspired by and extends the Euclidean mapping estimation framework~\cite{perrot2016mapping,seguy2018large} to the hyperbolic setting. 
Hyp-ME aims at finding a transport map between two hyperbolic 
representations using the GM as a proxy. 
We use a loss composed of two global terms: 
a fitting data term for the transport map and an OT-loss for the GM, 
each with its regularization term.
Thus, the total loss of the Hyp-ME is:
\begin{equation}
\label{eq:total_loss}
\begin{aligned}
g(\rmM,\, T) \coloneqq &
\Biggl(\underbrace{\recloss^\hyp(T,\, B_{\rmM}^\hyp) + \omega\, \Omega(T)}_{\text{Quality of the mapping}}\Biggr)
 + 
 \eta \Biggl(\underbrace{ \langle \rmM, \rmC\rangle_{\textF} - \epsilon\, H(\rmM)}_{\text{Regularized OT}}\Biggr),
\end{aligned}
\end{equation}
where $ \Omega(\cdot)$ is a regularization term and $\eta \geq 0,\, \omega  \geq 0$ are hyper-parameters.
In particular,  $\eta$ controls the trade-off between the two groups of terms in the loss.
We obtain $T$ and $\rmM$ by solving the following optimization problem
\begin{equation}
\label{eq:mapping_estimation}
(\rmM, T) \leftarrow \argmin_{T \in \gT,\,  \rmM \in \Pi(\rva, \rvb)} g(\rmM,\, T),
\end{equation}
where $\gT$ is the space of transformations from $\gX^\texts$ to $\gX^\textt$.
%

\paragraph{Optimization procedure.}
We rely on alternate optimization w.r.t. both parameters $\rmM$ and $T$. 
This algorithm well-known as Block Coordinate Descent~\cite{tseng2001convergence}.
Algorithm~\ref{alg:mapping_estimation} shows the steps of our optimization procedure. 
We note that the GM couples both groups of terms. 
Thus, minimizing the total loss with a fixed transport map boils down to a constrained OT-problem 
(see Eq.~\ref{eq:step_1}). 
We rely on a generalized conditional gradient descent algorithm (GCG)~\citep{courty2016optimal,rakotomamonjy2015generalized,bredies2005equivalence}.
We repeat alternating until we satisfy some convergence criterion. 
We are not insured to converge to the optimal solution as the data fit term is only convex when $s \rightarrow \infty$.
However, the algorithm works well in practice. 
%
%
\begin{algorithm}[h]
	\caption{Hyp-ME with Block Coordinate Descent}
	\label{alg:mapping_estimation}
	\begin{algorithmic}[1]
		\REQUIRE{Initial transport map $\hat{T}$, cost matrix $\rmC$} 
		\ENSURE{Stationary point of Eq.~\ref{eq:mapping_estimation}}
		\WHILE{not converge}
		\STATE \textbf{Update $\rmM$:} with a fixed estimator $\hat{T}$,
		\begin{equation}
		\label{eq:step_1}
		\hat{\rmM} \leftarrow \argmin_{\substack{\rmM \in \Pi(\rva, \rvb)}}  \eta \left(\langle \rmM, \rmC\rangle_{\textF} - \epsilon\, H(\rmM) \right)
		+ \recloss^\hyp(\hat{T},\, B_{\rmM}^\hyp).
		\end{equation}
		\STATE \textbf{Update $T$:} with a fixed coupling $\hat{\rmM}$, 
		\begin{equation}
		\label{eq:step_2}
		\hat{T} \leftarrow \argmin_{\substack{T \in \gT}}  
		\recloss^\hyp(T,\, B_{\hat{\rmM}}^\hyp) + \omega\, \Omega(T).
		\end{equation}
		\ENDWHILE
	\end{algorithmic}
\end{algorithm}
\paragraph{Updating $\rmM$.}
%
%
In our case, the GCG algorithm uses a partial linearization of the Hyp-ME loss 
(E.q.~\ref{eq:total_loss}) 
w.r.t. $\rmM$ to build a sequence of linear minimization problems.
Thus, for each iteration $k$, we compute the following linear oracle:
\begin{equation}
\label{eq:linear_oracle}
\begin{aligned}
\rmM^* 
\leftarrow \argmin_{\rmM \in \Pi(\rva, \rvb) } \left\langle \rmM,\,  \rmF^k \right\rangle_{\textF} -\hat{\epsilon}\, H(\rmM),
\end{aligned}
\end{equation}
where $\rmF^k = \eta\, \rmC + \left(\lambda_{\text{GM}}^s\right)^{-2} \nabla_{\rmM} \recloss^\hyp(\hat{T}, \, B_{\rmM}^\hyp)\biggr\rvert_{\rmM=\rmM^{k}}$, $\hat{\epsilon} = \eta\, \epsilon$, and 
$\lambda_{\text{GM}}^s$ is the conformal factor at the point mapped by the GM.
This factor links the hyperbolic and the Euclidean gradient (see Section~\ref{sec:hyperbolic_spaces}).
Eq.~\ref{eq:linear_oracle} is a regularized OT-problem with a modified cost function.
We update the map by computing $\rmM^{k+1}  = \rmM^{k} + \alpha^k \left( \rmM^* - \rmM^{k} \right)$, 
where $\alpha^k$ is the optimal step that satisfies the Armijo rule that 
minimizes Eq.~\ref{eq:total_loss}. 
We repeat until reaching some local convergence criterion.
\paragraph{Updating $T$.}
The Hyp-ME considers a broad set of transformations $\gT$. 
We use HNN. 
Thus, we rely on algorithms based on Riemannian gradient descent~\cite{becigneul2018riemannian}  to minimize Eq.~\ref{eq:step_2}, 
which is possible given that we have access to closed-form implementations of the 
gyrovector and Riemannian-based operations (see Section~\ref{sec:hyperbolic_spaces}). 
%

\subsection{Analysis: Hyperbolic Linear Layer}
In this section, we propose a novel analysis of the set of hyperbolic linear transformations induced by a real matrix $\rmL \in \sR^{\textd^\textt \times \textd^\texts}$:
\begin{equation}
\gT = \left\{ T: \exists \rmL \in \sR^{\textd^\textt \times \textd^\texts}, \, \forall \rvx^\texts \in \gX^\texts, \, T(\rvx^\texts) = \left(\rmL^{\otimes_s} \rvx^\texts \right)^\top  \right\}.
\end{equation}
Furthermore, we define $\Omega(T) = \|\rmL - \rmK \|_\textF^2$,  
where $\rmK$ is a predefined constant matrix, e.g., $\rmI$,
which ensures that the examples are not moved far away from their initial position.

\paragraph{Consistency.}
Showing the consistency boils down to bound with high probability the true risk $R(\rmL)$ by the empirical risk $\hat{R}_{\text{emp}}(\rmL)$, both defined as
\begin{equation*}
\begin{aligned}
R(\rmL) = \E_{(\rvx, \rvy)}\, d_s\left((\rmL^{\otimes_s} \rvx)^\top,\, \rvy\right)\quad \text{and} \quad
\hat{R}_{\text{emp}}(\rmL) = \frac{1}{n} \sum_{j=1}^{n} d_s\left((\rmL^{\otimes_s} \rvx_j)^\top,\, \rvy_j\right).
\end{aligned}
\end{equation*}
Note that the hyperbolic distance or its square is not geodesically convex w.r.t. 
M\"obius matrix multiplication (see Lemma~9 in supp.~mat.).

\begin{theorem}
	\label{thm:consistency}
	Let $\|\rvy\| \leq C_y$ and $\|\rvx\| \leq C_x$, for any $\rvx$ and $\rvy$ in $\sB_s^\textd\backslash \{\vzero\}$
	with a probability of $1 - \delta$  for any matrix $\rmL$ optimal solution of problem~\ref{eq:mapping_estimation} 
	such that $\|\rmL \rvx\|^{-1} \leq L$ and $\|\rmL \rvy\|^{-1} \leq L$, we have:
	\begin{equation}
	\label{eq:bounds_risk}
	\begin{split}
  R(\rmL) 
	 \leq & \hat{R}_{\text{emp}}(\rmL) + \frac{1}{\omega n}\left(\frac{N_x^2}{4\omega  n} + 2N_x + 8 K_y + 1\right) \\
	& + 
	\left(\frac{2}{ \omega }\left(\frac{N_x^2}{4\omega  n} + 2N_x + 1\right) + K_y\left(\frac{K_x + 16}{\omega}  + 1 \right)
	\right) \sqrt{\frac{\ln(1/\delta)}{2 n}},
	\end{split}
	\end{equation}
	where $N_x = \sqrt{8\pi s^3 LC_x}$, $K_x = d_s(\vzero, \, C_x)$, and $K_y = d_s(\vzero, \, C_y)$.
	%
	\begin{proof}
		We present the proof in supp.~mat.
	\end{proof}
\end{theorem}
Our theorem shows that the hyperbolic linear layer has a sample complexity 
of $O(n^{-1/2})$, the same as the Euclidean version~\citep{perrot2016mapping,perrot2015regressive}.
However, $s \rightarrow \infty$ does not link the bounds on the Euclidean and hyperbolic settings.

\paragraph{Quality of the approximation.}
We discuss the quality of the transport map approximation for the hyperbolic layer neural network.
Similarly to~\citep{perrot2016mapping}, we have the following theorem. 
\begin{theorem}
	Let $T^*$ be the true transport map.
	Let $B_{\rmM_0}^{\hyp}$ be the true GM associated with the probabilistic coupling $\rmM_0$.
	Let $B_{\hat{\rmM}}^{\hyp}$ be the empirical GM of $\rmX^s$ using the probabilistic 
	coupling $\hat{\rmM}$ learned between $\rmX^s$ and $\rmX^t$.
	Then, quality of the transport map approximation is bounded as follows
	\begin{equation}
	\label{eq:quality_of_transport_map_approx}
	\begin{split}
	 \E_{\rvx^\texts \sim \gX^\texts}\left[d_s\left(T(\rvx^\texts), \, T^*(\rvx^\texts)\right)\right] 
	\leq & \sum_{\rvx^\texts \in \gX^\texts} d_s\left(T(\rvx^\texts), \, B_{\hat{\rmM}}^{\hyp}(\rvx^\texts)\right)  \\
	&+ \sum_{\rvx^\texts \in \gX^\texts} d_s\left(B_{\hat{\rmM}}^{\hyp}(\rvx^\texts), \, B_{\rmM_0}^{\hyp}(\rvx^\texts)\right)  \\
	& +  \E_{\rvx^\texts \sim \gX^\texts}\left[d_s\left(T^*(\rvx^\texts),\, B_{\rmM_0}^{\hyp}(\rvx^\texts)\right)\right]
	+ O\left(\frac{1}{\sqrt{n}}\right).
	\end{split}
	\end{equation}
	%
	\begin{proof}
		We present the proof in supp.~mat.
	\end{proof}
\end{theorem}
The inequality in Eq.~\ref{eq:quality_of_transport_map_approx} assesses the 
quality of the learned transformation $T$ w.r.t three principal quantities. 
We minimize the first quantity in Eq.~\ref{eq:mapping_estimation}.
We foresee the first term to be small for families of functions with 
high representation power, e.g., deep HNN; however, we need to explore these families' consistency. 
The remaining terms are theoretical quantities that are hard to bound due to GM's lack of theory in the statistical learning context. 
However, we expect the second term to decrease uniformly w.r.t. the number of samples as it measures the empirical GM's quality. 
Assuming that the empirical GM is a good approximation, we also expect the last term to be small. 
We believe understanding gyrobarycenters opens up new theoretical perspectives about OT in machine learning on hyperbolic manifolds, but these are beyond the scope of this paper.

\paragraph{Link with wrapped Gaussians.}
This section shows the link between the hyperbolic linear layer and 
wrapped Gaussian distributions in the hyperbolic space. 
We start by defining wrapped Gaussian distributions. 
Then, we state a theorem that links both models explicitly.
\begin{definition}[Wrapped Gaussian~\citep{nagano2019wrapped, mardia2009directional,said2019warped}\label{def:wrapped_gaussian}] 
	Define $\rvx \sim \left(\Exp_{\mu}\right)_{\sharp} \gN(0, \Sigma)$ to be a wrapped Gaussian random variable with
	bias $\mu \in \sB_s^\textd$ and covariance matrix $\Sigma \in \sR^{n\times n}$.
	To build a wrapped Gaussian, we draw samples at random from a zero-mean Gaussian distribution 
	with covariance $\Sigma$, $\rvz \sim \gN(0, \Sigma)$.
	We project these samples onto the manifold at zero, $\bar{\rvx}_i = \Exp_{\vzero}(\rvz_i)$. 
	Finally, we add the bias term $\mu$ using the M\"obius addition,  $\rvx_i = \Exp_{\mu}(\rvz_i)$.
\end{definition}

\begin{theorem}
	\label{thm:wrapped_linear_model}
	Let $\rvx \sim \left(\Exp_{\mu_1}\right)_{\sharp} \gN(0, \Sigma_1)$ and 
	$\rvy \sim \left(\Exp_{\mu_2}\right)_{\sharp} \gN(0, \Sigma_2)$ be two hyperbolic random variables, distributed under wrapped Gaussian 
	with parameters  $\mu_i \in \sB_s^\textd$ for $i \in [2]$ and $\Sigma_i \in \sR^{n \times n}$ for $i \in [2]$. 
	Then,  
	\begin{equation}
	\label{eq:wrapped_gaussians}
	\begin{bmatrix}
	\rvx\\
	\rvy
	\end{bmatrix}
	\sim \left(\Exp_{[\mu_1, \mu_2]}\right)_{\sharp}
	\gN \left(
	\vzero, 
	\begin{bmatrix}
	\Sigma_1 &  \rmT\, \Sigma_1\\
	\Sigma_1\, \rmT & \Sigma_2\\
	\end{bmatrix}
	\right),
	\end{equation}
	where 
	%
	$\rmT = \Sigma_1^{-\frac{1}{2}} \left(\Sigma_1^{\frac{1}{2}} \Sigma_2 \Sigma_1^{\frac{1}{2}} \right)^{\frac{1}{2}} \Sigma_1^{-\frac{1}{2}}$~\citep{bhatia2019bures,flamary2019concentration}.
	Then, the transport map is $T(\rvx) = \rvy  = \mu_2 \oplus_s \rmT^{\otimes_s} (-\mu_1 \oplus_s \rvx)$.
	\vspace{-8pt}
	\begin{proof}
		We present the proof in supp.~mat.
	\end{proof}
\end{theorem}
Observe that $\lim_{s \rightarrow \infty} T(\rvx) = \mu_2 + \rmT\, (\rvx - \mu_1)$, 
which is the transport map between two Gaussians~\cite{knott1984optimal,peyre2019computational}.
%
%

The main implication of Theorem~\ref{thm:wrapped_linear_model} is that by definition (see Eq.\ref{eq:hyperbolic_neural_network}), a hyperbolic linear layer
represents a linear Monge mapping between wrapped Gaussian distributions.
%

\begin{figure}[h]
	\centering
	{\includegraphics[width=.4\linewidth, trim={0 0mm 0mm 0mm}, clip]{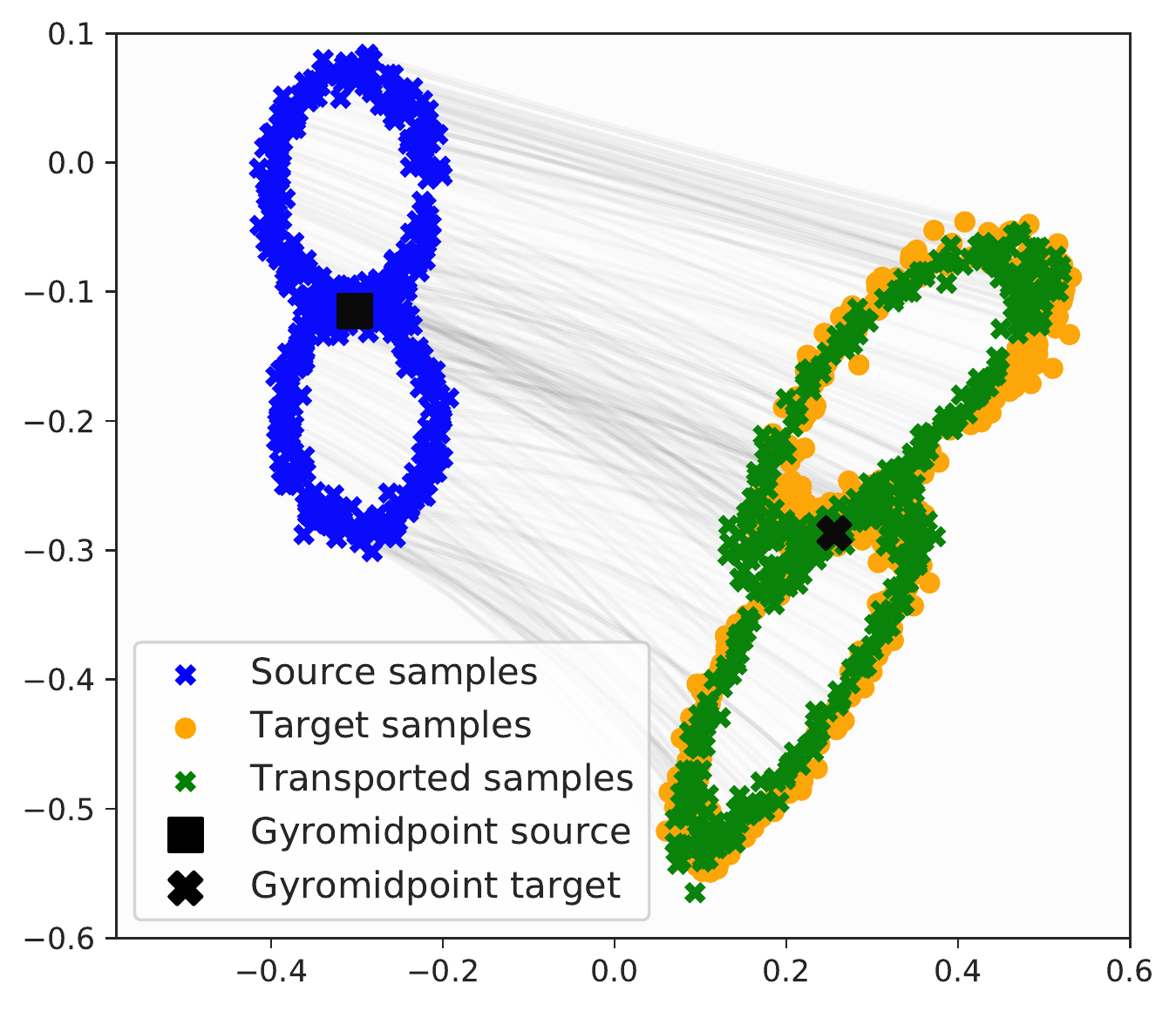}}	
	\caption{
		\small
		\textbf{Hyperbolic mapping estimation on a toy example.} 
		Example of wrapped linear mapping estimation between empirical distributions.
		Lines denote the geodesics (gyrolines) of the transported samples.
	}
	\label{fig:toy_example}
\end{figure}

\paragraph{The Wrapped Linear Map (W-linear map).}
It is the closed-form solution given by the Theorem~\ref{thm:wrapped_linear_model}.
First, we center the data by subtracting (i.e., left M\"obius subtraction) the gyromidpoint of source and target, respectively.
Then, we compute the Gaussian transport plan~\citep{bhatia2019bures,masarotto2019procrustes} 
of the covariance matrices in the tangent space at zero, $\rmT$.  
Fig.~\ref{fig:toy_example} shows an illustration of the wrapped linear transport.
%

\section{Experiments}
\subsection{Setup.~\label{sec:setup}}
\paragraph{Datasets.}
We consider two subtasks of the OAEI 2018 ontology matching challenge~\cite{algergawy2018results}: 
\small{\emph{Anatomy}}, with two ontologies; \small{\emph{Biodiv}}, 
with four\footnote{oaei.ontologymatching.org/2018/}.
We use the \small{\emph{English}-\emph{French}}-matching task from the DBPL15k dataset~\cite{sun2017cross},  
used for cross-lingual entity alignment (see Section 1 in the supp.~mat.).
For each collection, we generate embeddings in the Poincar\'e Ball of dimension $10$~\cite{nickel2017poincare}.
We perform a 10-fold cross-validation loop.
We consider a supervised aligning task, where we train with $10\%$ of the matched data
and validate on the remaining $90\%$.

\paragraph{Baselines and metrics.}
We compare several OT-based alignment methods against their proposed hyperbolic counterpart:
Linear mapping estimation~\citep{knott1984optimal}, 
mapping estimation (ME)~\cite{perrot2016mapping}, 
OT-DA~\citep{courty2016optimal}, 
hyperbolic OT-direct~\citep{alvarez2019unsupervised}, 
Euclidean OT-direct, 
hyperbolic linear layer (W-linear map), and
hyperbolic OT-DA, and Hyp-ME.
All models return transported embeddings. 
Using these, we retrieve nearest neighbors under the hyperbolic distance. 
We report the Hits$@10$, which measures the percentage of times the right matched point appears in the top $10$ ranked evaluation.
Additionally, we report the computation time.  

We use neural networks and HNN for the Euclidean and hyperbolic cases, respectively.
We set 6 hidden layers, 100 hidden units with ReLU as the activation function.
We initialize the weights as in~\citep{glorot2010understanding}.
We use Riemannian ADAM~\citep{becigneul2018riemannian} for hyperbolic models, and ADAM~\cite{kingma2014adam} for the Euclidean ones.
In all experiment, we set the number of iterations of the Sinkhorn algorithm to be less than $100$.
We set the regularization parameter $\epsilon=0.01$, and $\rva = \rvb = \vone_{n}/ n$. 
For iterative methods, we set the stopping criteria to a relative tolerance of $1\times 10^{-7}$.

\paragraph{Technical aspects.}
We used Gensim~\cite{rehurek_lrec} to compute the Poincar\'e embeddings. 
We used the POT library~\cite{flamary2017pot} for OT related algorithms. 
We used the Autodiff implementation of the Sinkhorn algorithm\footnote{
	\url{github.com/gpeyre/SinkhornAutoDiff}
}
for the end-to-end learning task.
We relied on the geoopt package for Riemannian optimization~\citep{geoopt}.
Our methods are implemented in Pytorch~\cite{NEURIPS2019_9015} and are available\footnote{
	\href{https://github.com/ahoyosid/hyperbolic_alignment}{Link to the Github repository}
}.
We run all the experiments on a single desktop machine with a 4-core Intel Core i7 $@2.4$ GHz.

\subsection{Results}
\begin{figure}[h]
	\centering
	\includegraphics[width=.65\linewidth, trim={0 0mm 0mm 0mm}, clip]{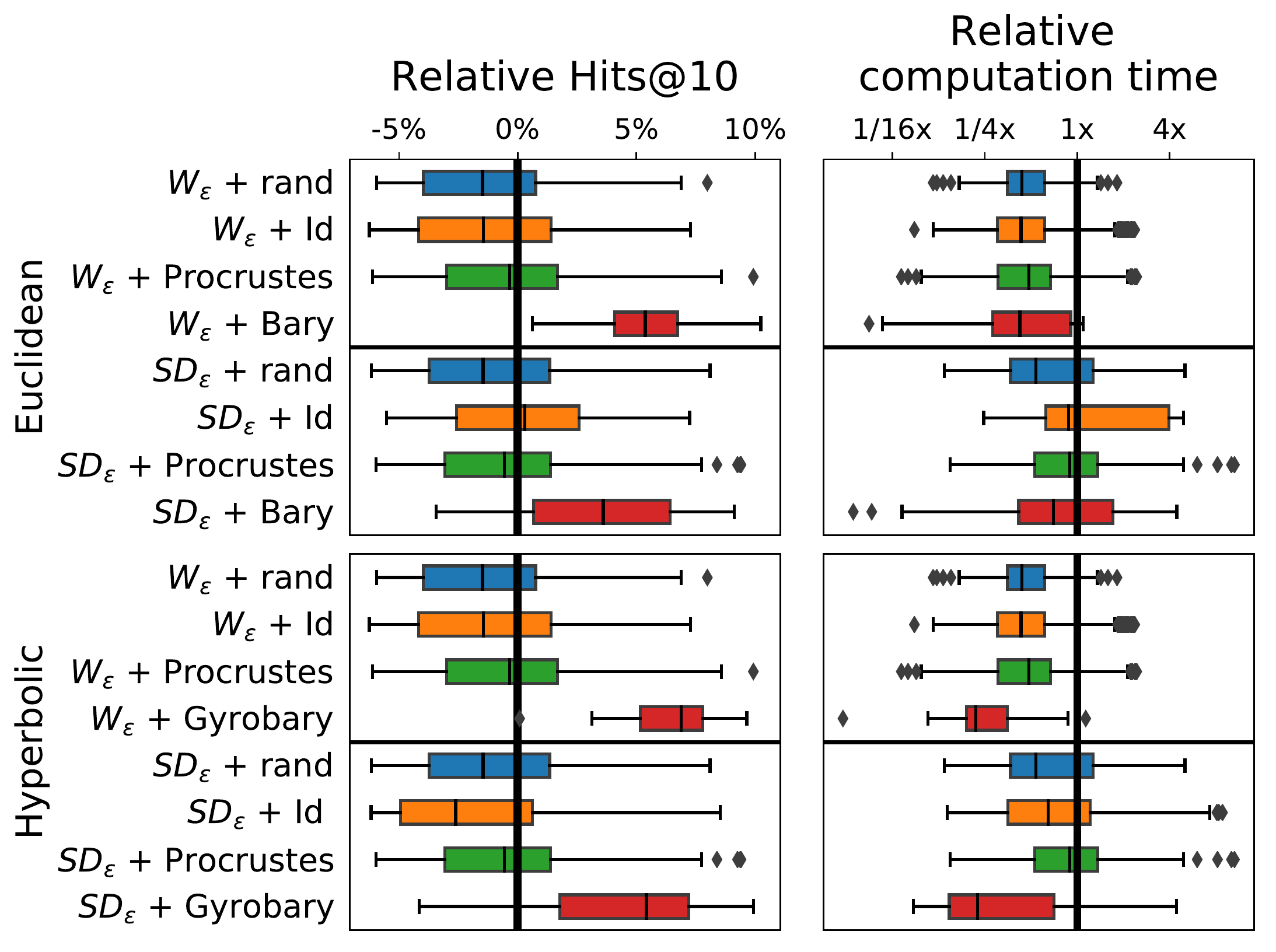}
	\caption{\small \textbf{Effect of various initialization in OT-direct:} 
		Comparing different initialization approaches for Euclidean and hyperbolic estimators, 
		using the  Sinkhorn loss and Sinkhorn divergence.
	}
	\label{fig:benchmark_init}
\end{figure}
\paragraph{Effect of the initialization in OT-direct.}
%
We explore the impact of several initialization strategies in the robustness of OT-direct: 
random initialization, identity mapping (Eq.~\ref{eq:identity}), 
Procrustes alignment (Eq.~\ref{eq:procustes}), 
and barycenter/gyrobarycenter mapping (Eq.~\ref{eq:gm_initialization}).
We use $W_\epsilon$ and $SD_\epsilon$ as OT-based losses.
The GM initialization corresponds to one step of the Algorithm~\ref{alg:mapping_estimation}
without taking the fitting data term ($\eta \rightarrow \infty$). 

Fig.~\ref{fig:benchmark_init} displays the performance across datasets relative to the mean of both 
Euclidean and hyperbolic neural networks with different initialization methods with two OT-direct losses:
$W_\epsilon$ and $SD_\epsilon$.
The barycenter and gyrobarycenter initialization display the best performance for all losses consistently. 
The random and the identity strategy exhibit similar performance in Euclidean and hyperbolic methods for all losses. 
The gyrobarycenter with the Sinkhorn loss has the best performance with less variance. 
Regarding the computation time, the gyrobarycenter is approximately two times faster than other initialization procedures. 
In Euclidean spaces, the barycenter method improves retrieval performance without increasing computation time. 
Therefore, the barycenter mapping/GM initialization provides a good proxy for the transport map. 
Hence, closer to the optimal value.
\paragraph{Retrieval.}
\begin{table}
	\resizebox{\textwidth}{!}{%
		\begin{tabular}{@{}l*{14}{c}c@{}}
			\toprule
			%
			&   \multicolumn{3}{c}{En-Fr} 	&&	\multicolumn{3}{c}{Human-Mouse}    &&    \multicolumn{3}{c}{Flopo-Pto} && \multicolumn{3}{c}{Envo-Sweet} \\
			\cline{2-4} \cline{6-8} \cline{10-12} \cline{14-16} \noalign{\smallskip}
			&\multirow{2}{*}{T(m)} &  \multicolumn{2}{c}{Hits@10}   && \multirow{2}{*}{T(m)} & \multicolumn{2}{c}{Hits@10} && \multirow{2}{*}{T(m)} &\multicolumn{2}{c}{Hits@10} && \multirow{2}{*}{T(m)}&\multicolumn{2}{c}{Hits@10}\\
			&&$\rightarrow$&$\leftarrow$&&&$\rightarrow$&$\leftarrow$&&&$\rightarrow$&$\leftarrow$&&&$\rightarrow$&$\leftarrow$\\
			\midrule
			%
			\emph{Euclidean Methods}  &&&&&&&&&&&&&&&\\
			\noalign{\smallskip}
			\lvl Linear ma~\citep{knott1984optimal}& $< 0.1$ &\color{grays}{\round{2}{0.473177}}&\color{grays}{\round{2}{0.473177}}&&
			$< 0.1$&\color{grays}{\round{2}{0.586081}}&\color{grays}{\round{2}{0.673993}}&&
			$< 0.1$&$\bm{8.77}$&\round{2}{6.811594}&&
			$< 0.1$&\color{grays}{\round{2}{2.853186}}&\color{grays}{\round{2}{2.770083}}\\
			\lvl OT-DA~\citep{courty2014domain}   & $< 0.1$ & \round{2}{14.930002}&\round{2}{14.943305}&&
			$< 0.1$&$\bm{13.15}$&\round{2}{12.256410}&&
			$< 0.1$&\round{2}{7.028986}&\round{2}{6.956522}&&
			$< 0.1$&\color{grayl}{\round{2}{13.961219}}&\round{2}{14.376731}\\
			\lvl ME~\citep{perrot2016mapping} & \round{2}{1.787282} &\round{2}{14.859049} & \round{2}{14.878228}&&
			\round{2}{1.094943}&\round{2}{12.461538}&\color{grayl}{\round{2}{9.421245}}&&
			\round{2}{0.130789}&\color{grayl}{\round{2}{6.884058}}&\round{2}{6.956522}&&
			\round{2}{0.387478}&\round{2}{14.072022}&\round{2}{14.155125}\\
			
			\noalign{\smallskip}
			\lvl\emph{OT-Direct} \quad\,  &&&&&&&&&&&&&&&\\
			\lvl\lvl $W_\eps$ + Bary    & \round{2}{0.709181}& \color{grays}{\round{2}{3.611841}}&\color{grays}{\round{2}{3.495199}} &&
			\round{2}{3.951266}&\color{grayl}{\round{2}{11.179487}}&\color{graym}{\round{2}{8.249084}}&&
			\round{2}{0.817022}&\round{2}{6.811594}&\round{2}{6.159420}&&
			\round{2}{1.030887}&\color{grayl}{\round{2}{13.767313}}&\color{grayl}{\round{2}{12.880886}}\\	
			\lvl\lvl$SD_\eps$ + Bary    & \round{2}{0.942067}& \color{grays}{\round{2}{3.484414}}& \color{grays}{\round{2}{3.500720}} &&
			\round{2}{11.380251}&\color{grayl}{\round{2}{11.589744}}&\color{graym}{\round{2}{7.956044}}&&
			\round{2}{1.498617}&\color{grayl}{\round{2}{6.521739}}&\round{2}{6.340580}&&
			\round{2}{2.912920}&\color{grayl}{\round{2}{13.933518}}&\color{grayl}{\round{2}{13.213296}}\\						
			\midrule
			\noalign{\smallskip}
			\emph{Hyperbolic Methods}  &&&&&&&&&&&&&&&\\
			\noalign{\smallskip}
			\lvl W-linear map  & $< 0.1$ &\color{grays}{\round{2}{0.359826}}& \color{grays}{\round{2}{0.359826}}&&
			$< 0.1$&\color{grays}{\round{2}{0.586081}}&\color{grays}{\round{2}{0.761905}}&&
			$< 0.1$&\color{grayl}{\round{2}{6.884058}}&\round{2}{6.956522}&&
			$< 0.1$&\color{grays}{\round{2}{2.520776}}&\color{grays}{\round{2}{2.603878}}\\
			\lvl OT-DA    & $< 0.1$ & \round{2}{14.943305}& \round{2}{14.930002}&&
			$< 0.1$&\round{2}{12.967033}&$\bm{12.80}$&&
			$< 0.1$&{\round{2}{6.956522}}&\round{2}{6.956522}&&
			$< 0.1$&\round{2}{14.515235}&$\bm{14.93}$\\
			\lvl ME   & \round{2}{25.592695} & $\bm{16.17}$ & $\bm{15.87}$&&
			\round{2}{37.146387}&\round{2}{13.032967}&\round{2}{11.465201}&&
			\round{2}{2.393552}&\round{2}{7.246377}&\round{2}{6.666667}&&
			\round{2}{9.916532}&$\bm{14.57}$&\color{grayl}{\round{2}{13.988920}}\\
			\noalign{\smallskip}
			\lvl\emph{OT-Direct}~\citep{alvarez2019unsupervised} \quad\,  &&&&&&&&&&&&&&&\\
			\lvl\lvl$W_\eps$ + Gyrobary   & \round{2}{5.391374}& \color{grays}{\round{2}{3.870859} }&\color{grays}{\round{2}{3.915627} }&&
			\round{2}{1.153478}&\round{2}{12.644689}&\color{grayl}{\round{2}{9.919414}}&&
			\round{2}{0.537575} &\color{graym}{\round{2}{6.014493}}&\round{2}{6.376812}&&
			\round{2}{1.596191}&\round{2}{14.321330}&\color{grayl}{\round{2}{13.988920}}\\	
			\lvl\lvl$SD_\eps$ + Gyrobary   & \round{2}{6.183168}& \color{grays}{\round{2}{3.908870}}& \color{grays}{\round{2}{3.911478} }&&
			\round{2}{3.803670}&\round{2}{13.040293}&\color{grayl}{\round{2}{10.336996}}&&
			\round{2}{0.562992}&\round{2}{7.101449}&\round{2}{6.594203}&&
			\round{2}{2.716516}&\round{2}{14.560}&\color{grayl}{\round{2}{13.628809}}\\					
			\bottomrule
	\end{tabular}}
	\caption{\small Prediction on DBP15K, Anatomy, and Biodiv datasets.
		Color-code: the darker the better.}
	\label{tab:tax_table}
\end{table}
Table~\ref{tab:tax_table} presents the predictive performance for the supervised alignment task. 
The hyperbolic version of OT-DA and ME display the best performance across datasets. 
Their Euclidean version follows them with similar performance. 
In the Flopo-Pto subtask of the Biodiv dataset, the Poincar\'e embeddings exhibit a Gaussian-like form, explaining linear mappings' good performance. 
We do not observe this in other datasets. OT-direct approaches behave poorly in the DBP15K dataset. 
The input vectors lie inside the unit ball; thus, the Euclidean barycenter will always be in this ball. Additionally, we use estimators with a high representation power. 
Therefore, we attribute the slight difference in performance between the hyperbolic and Euclidean schemes to the quality of the probabilistic coupling $\rmM$ on each manifold.

\section{Discussion}
In this paper, we show that the hyperbolic layer has a sample complexity of $O(n^{-1/2})$
(Theorem~\ref{thm:consistency}).
We then bridge the gap between the hyperbolic layer and wrapped Gaussian distributions using Theorem~\ref{thm:wrapped_linear_model}.
Additionally, we proposed a closed-form solution for the transport between wrapped Gaussian distributions: the wrapped linear model.
We extend to the hyperbolic space some OT-based methods for feature alignment by using gyrobarycenter mappings: Hyp-OT-DA and Hyp-ME. 

In all experiments, the hyperbolic methods benefit the retrieval performance compared to the Euclidean geometry. 
The barycenter and gyrobarycenter initialization approaches help to avoid the estimator's collapse while minimizing OT-based similarities directly.
The wrapped linear model can align the latent space of hyperbolic wrapped variational autoencoders~\citep{mathieu2019continuous}.
The analysis presented in this paper opens the door to several perspectives on machine learning on hyperbolic spaces, such as the quality 
and the consistency of the gyrobarycenter approximation in OT problems in hyperbolic spaces.

\bibliographystyle{abbrvnat}
\bibliography{biblio.bib}

\newpage

\appendix
\setcounter{theorem}{0}
\setcounter{lemma}{0}
\input{supp_mat.tex}

\end{document}

%% file: supp_mat.tex
\subsection{Data description}

\paragraph{DBP15K~\cite{sun2017cross}.}
It is a subtask from DBpedia, which is a large-scale multilingual knowledge base that includes 
inter-language links (ILLs) from entities of English version to those in other languages.
The DBP15K dataset consists of 15 thousand ILLs with popular entities from English to Chinese, Japanese, and French respectively.

\setlength{\tabcolsep}{5pt}
\begin{table*}[h]
	\centering
	\footnotesize
	\resizebox{\textwidth}{!}{%
		\begin{tabular}{r*{9}{c}r}
			\toprule
			&    \multicolumn{2}{c}{DBP15K} 	&&	\multicolumn{2}{c}{Anatomy}    &&    \multicolumn{4}{c}{Biodiv}					\\
			\cline{2-3} \cline{5-6}  \cline{8-11} \noalign{\smallskip}
			& English (EN) & French (FR) &&		Human & Mouse		&&	FLOPO &	PTO	& ENVO	& SWEET	  \\	
			\midrule
			Entities    & 19993 & 19661&& 3298	&	2737    && 360  &  1456   & 6461     &4365  	\\
			Relations & 115722 & 105998 &&	18556   &  7364   && 472   &  11283  &73881   &30101	\\
			Matched & \multicolumn{2}{c}{14278} &&  \multicolumn{2}{c}{1517}  && \multicolumn{2}{c}{154} & \multicolumn{2}{c}{402}\\     
			\bottomrule
	\end{tabular}}
			\caption{\small Dataset characteristics.}
\end{table*}

\subsection{Additional results}

\begin{figure}[h]
	\small
	\centering
	%
	%
	\subfloat[][Initial classifer. \\ Accuracy: 0.57]{
		\includegraphics[width=0.33\linewidth, trim={0 0mm 0mm 0mm}, clip]{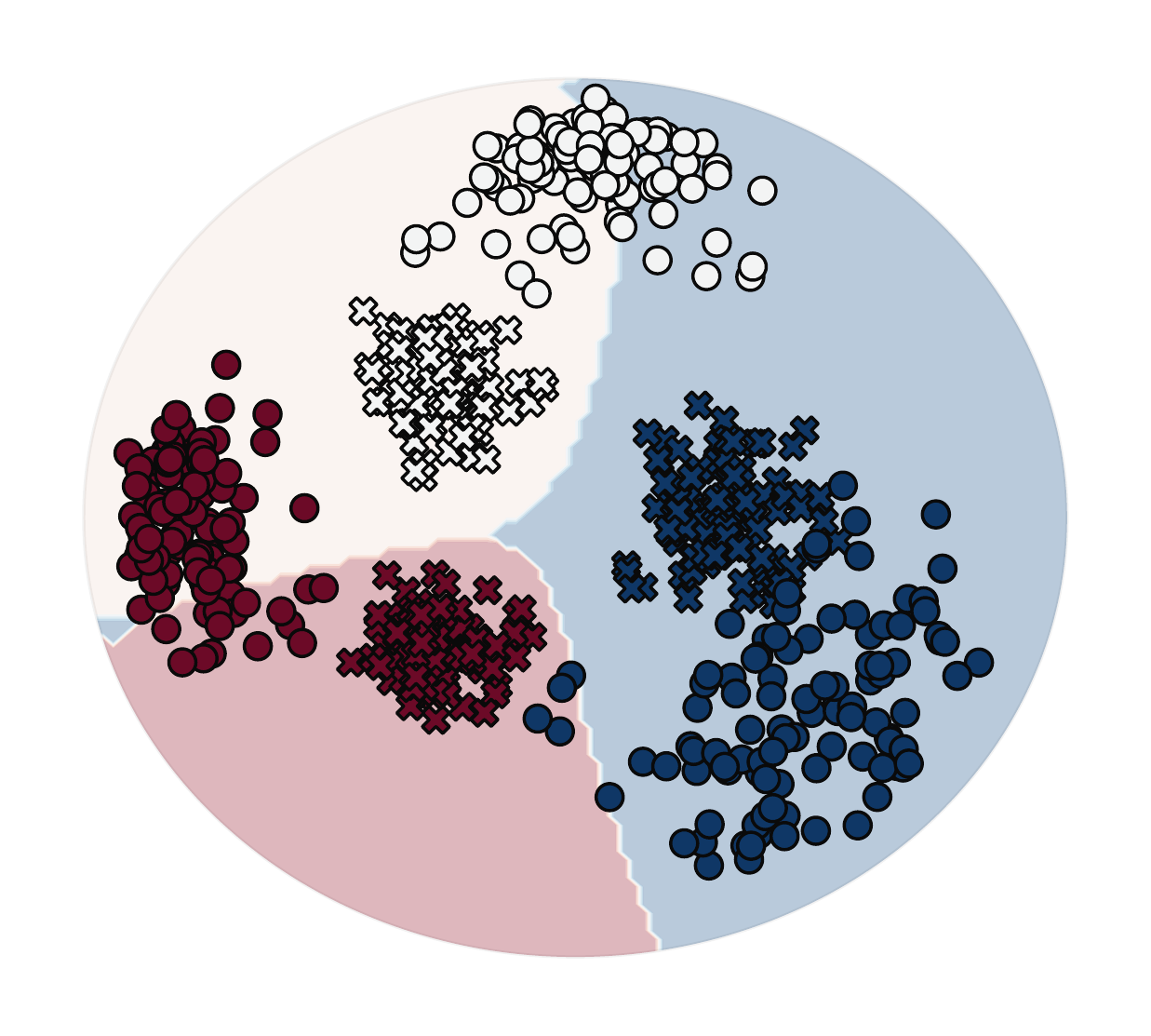}
	}	
	\subfloat[Gyrolines]{\includegraphics[width=0.33\linewidth, trim={0 0mm 0mm 0mm}, clip]{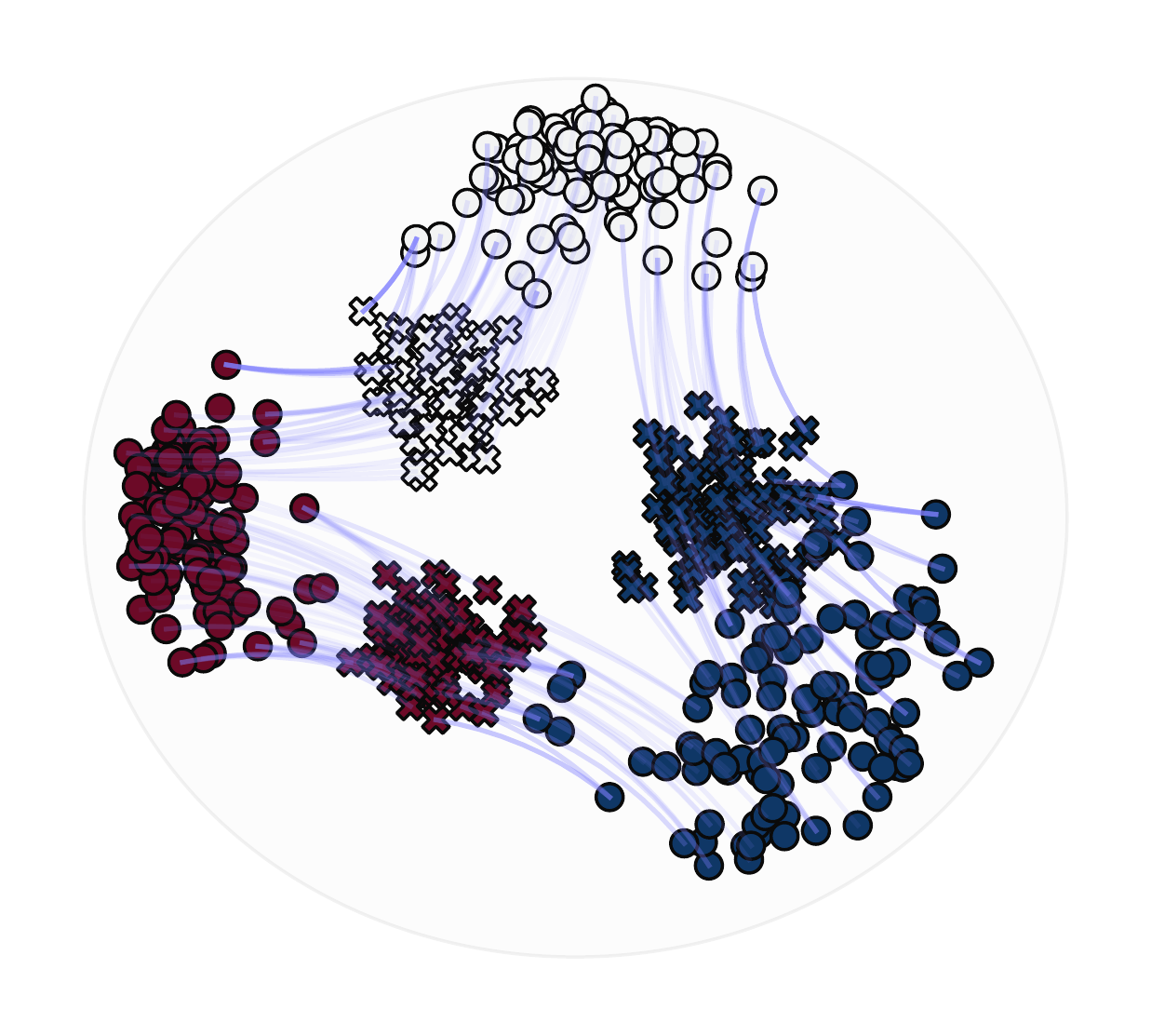}}	
	\subfloat[][After fine-tuning. \\ Accuracy: 0.86]{\includegraphics[width=0.33\linewidth, trim={0 0mm 0mm 0mm}, clip]{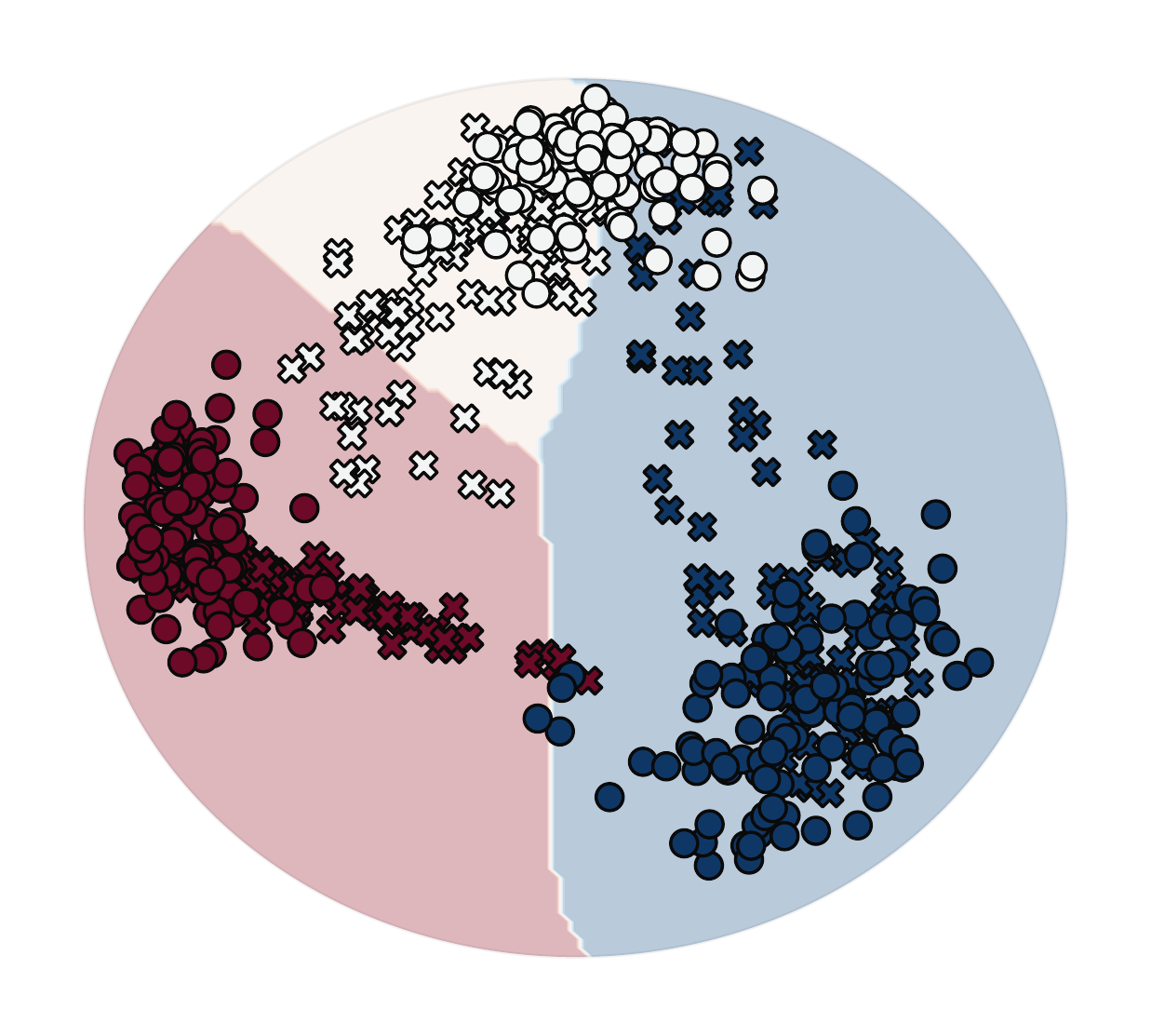}}	
	\caption{
		\small
		\textbf{Illustration on a toy example:} 
		(a)~Decision boundaries for shallow hyperbolic network trained on 
		the source domain.
		(b)~Gyrolines between samples matched according to the transport plan.
		(c)~Decision boundaries after fine-tuning the classifier with samples transported to the target domain.
		Crosses depict the source domain, while class-colored circles are the target domain samples.
	}
	\label{fig:example_domain_adaptation}
\end{figure}

\paragraph{Domain adaptation on synthetic data.}
The aim of domain adaptation is to build a classifier such that said classifier trained on a domain (source) 
can be used to predict in a different domain (target).
Fig.~\ref{fig:example_domain_adaptation} shows an illustration of domain adaptation by aligning 
the hyperbolic embeddings of both domains.

\begin{figure*}[h]
	\small
	\centering
	\begin{tabular}[c]{cc}
		\subfloat[][]{
			\includegraphics[width=0.2\linewidth, trim={0 0mm 0mm 0mm}, clip]{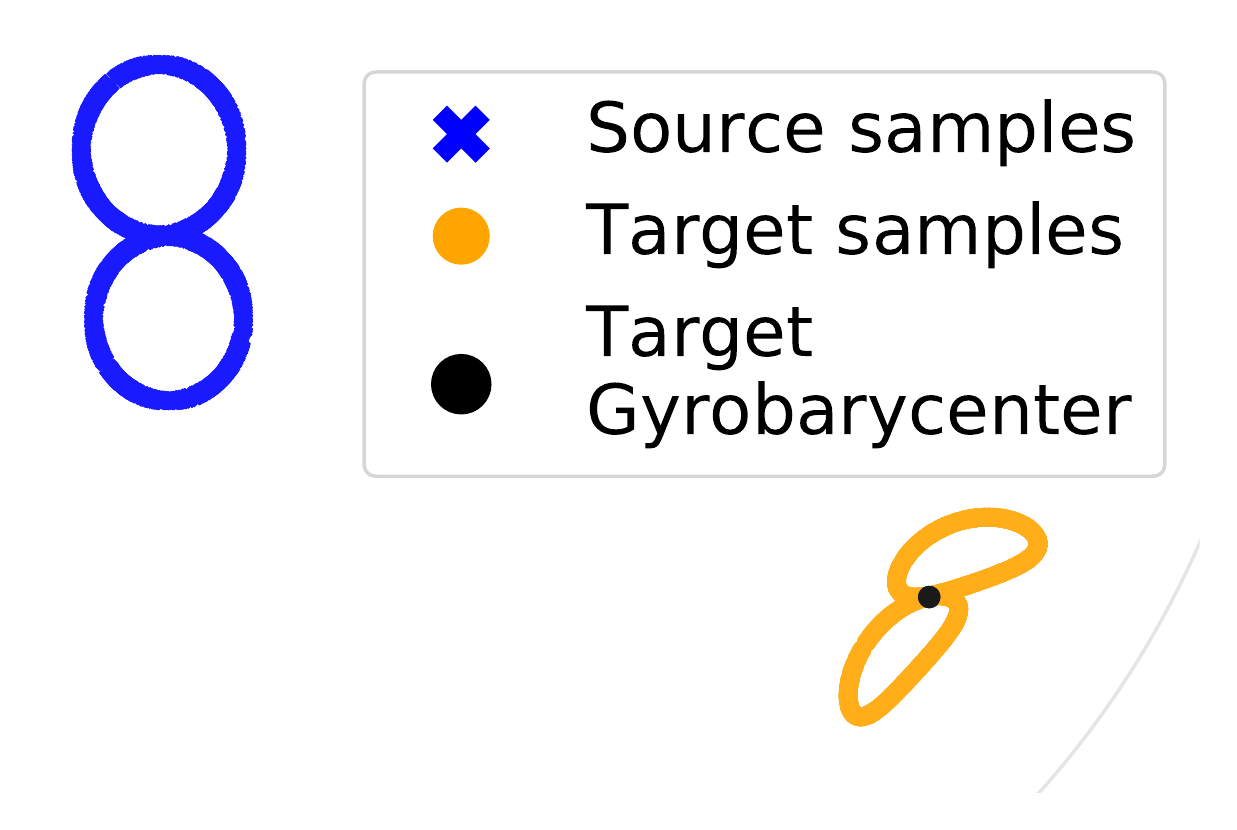}}
		&
		\subfloat[][Barycenter projection.]{\includegraphics[width=0.75\linewidth, trim={0 0mm 0mm 0mm}, clip]{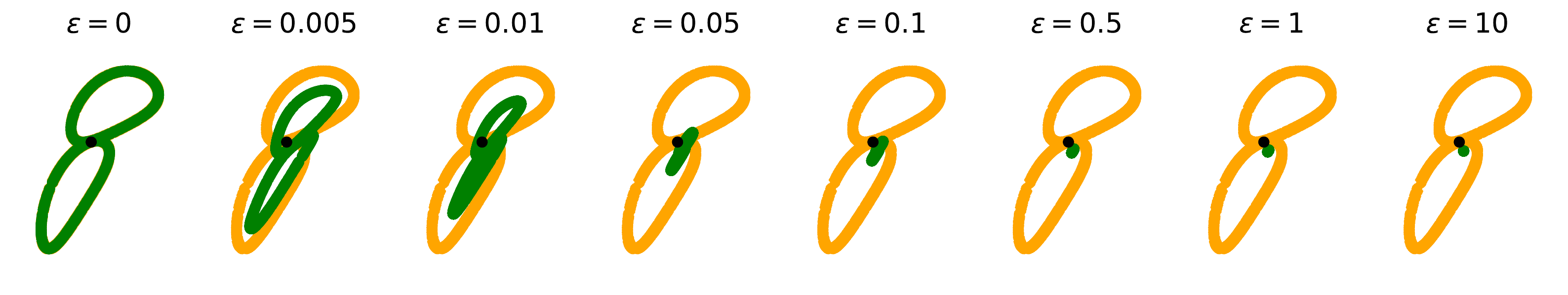}}\\
		\subfloat[][\\ Procrustes]{
			\includegraphics[width=0.12\linewidth, trim={0 0mm 0mm 0mm}, clip]{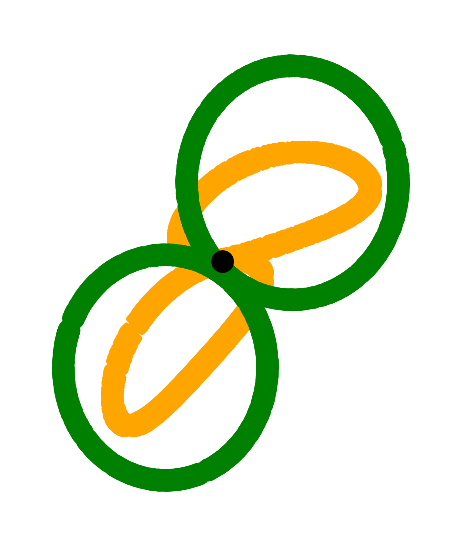}}
		& \subfloat[][Gyrobarycenter projection.]{
			\includegraphics[width=0.75\linewidth, trim={0 0mm 0mm 0mm}, clip]{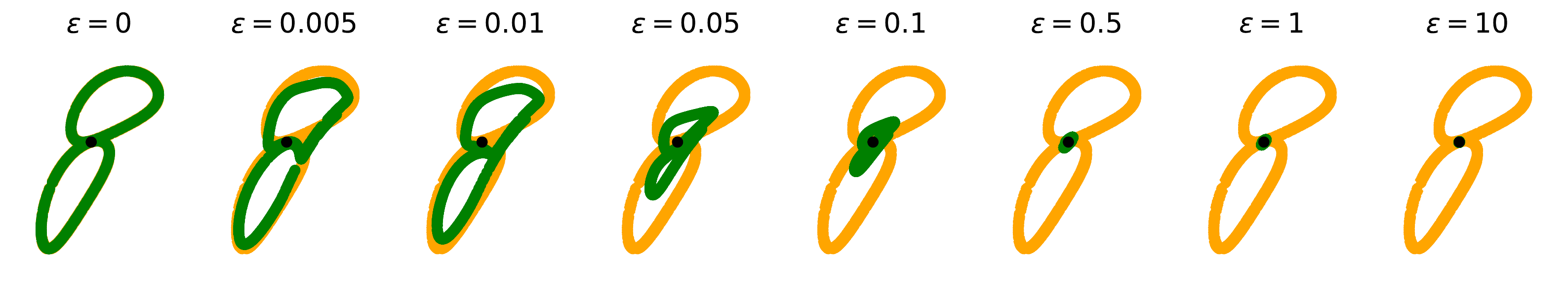}}	
	\end{tabular}
	\caption{
		\small
		\textbf{Illustration of barycenter and gyrobarycenter mappings:}
		(a) Original samples;
		(b) Euclidean Barycenter mapping;
		(c) Orthogonal Procrustes with data centering;
		(c) Hyperbolic barycenter mapping (gyrobarycenter).
		The gyrobarycenter mapping collapses into gyromidpoint for a large enough entropic regularization parameter.
	}
\end{figure*}

\subsection{Optimal transport on Riemannian manifolds\label{app:transport_riemann}}

\paragraph{Brenier and McCann mappings.}
Brenier's theorem states that the solution of the Monge problem can be characterized 
by the existence of the gradient of a convex function.
McCann's theorem \citep{mccann2011five, mccann2001polar} generalizes the 
Brenier's theorem to a general compact Riemannian manifold, $(\gM,\, \rho)$.
Let $\mu$ and $\nu$ be two probability measures compactly supported on $\gM$, where 
$\mu$ is absolutely continuous with respect to the Riemannian volume.  
According to McCann, the optimal transport map $T$ for 
the quadratic cost $c(\rvx, \rvy) = \frac{1}{2}d_s(\rvx, \rvy)^2$ can be written as 
$T(\rvx)= \Exp_\rvx\left(-\nabla \psi(\rvx)\right)$,
for some function $\psi$ such that $\psi^{cc} = \psi$, 
where $u^{c} = \inf_{\rvx \in \gX}\left\{c(\rvx, \rvy) - u(x)\right\}$ denotes the $c$-transform of $u$. 
A map $u$ such that $u^{cc} = u$ is called $c$-concave.

\paragraph{Some conditions for Regularity.}
\begin{wraptable}{r}{0.37\linewidth}
	\vspace{-10pt}
	\centering
	\small
	\begin{tabular}{rc}
		\toprule
		\multirow{2}{*}{cost function} & MTW  \\
		& condition\\
		\midrule
		$-\cosh \circ d$ & strong \\
		$-\log \circ (1 + \cosh) \circ d$ & strong \\
		$\pm\log \circ \cosh \circ d$ & weak \\
		\bottomrule
	\end{tabular}
	\caption{Hyperbolic cost functions that satisfy the strong/weak MTW conditions of regularity.}
	\label{table:hyperbolic_cost_functions}
\end{wraptable}
Ma-Trudinger-Wang (MTW) conditions for regularity~\citep{ma2005regularity, trudinger2009second} are sufficient for the optimal map to be smooth between a pair of log smooth bounded probability densities 
\citep{mccann2011five}. 
However, to use the squared-Riemannian distance of the hyperbolic space~\citep{li2009smooth, lee2009new} violates one of the MTW conditions as 
significant cut-locus issues arise in this context.
Nevertheless, we can satisfy the MTW conditions for cost functions which are composition of a function $l$ with a Riemannian distance of constant sectional curvature~\citep{lee2009new}.
The Table~\ref{table:hyperbolic_cost_functions} summarizes the MTW-regular cost functions.

\subsection{Properties of M\"obius operations~\label{sec:properties}}
\paragraph{Equivalent hyperbolic distances.}
\begin{equation}
\label{eq:hyperbolic_distances}
\begin{split}
d_s(\rvx, \rvy) &=  
2 s\, \tanh ^{-1}\left(\frac{\left\|-\rvx \oplus_{s} \rvy\right\|}{s}\right)\\
&= 2s\, \sinh^{-1}\left(\gamma_{\rvx}\gamma_{\rvy}\frac{\|\rvx - \rvy \|}{s}\right) \quad \text{(Theorem 7.4~\cite{beardon2007hyperbolic})}\\
\end{split}
\end{equation}
We have $\lim_{s \rightarrow \infty} d_s(\rvx, \rvy)  = 2\,  d(\rvx, \rvy)$~\citep{ungar2008analytic}.
In particular, if we compute the distance to $\vzero$~\citep{ratcliffe2006foundations, kim2013unit}, it reduces to 
\begin{equation}
\label{eq:hyperbolic_distace_to_zero}
\begin{split}
d_s(\vzero, \rvw) &= 
2 s\,  \tanh^{-1}\left(\frac{\|\rvw\|}{s}\right)\\
& = 2s\, \sinh^{-1}\left(\gamma_\rvw \frac{\|\rvw\|}{s}\right). \\
\end{split}
\end{equation}

\paragraph{The M\"{o}bius scalar multiplication.} 
It is defined as 
\begin{equation}
\label{eq:mobius_scalar_multiplication}
\begin{split}
r \otimes_{s} \rvx  =  &  s\, \tanh 
\left(r \tanh ^{-1}\left(\frac{\|\rvx\|}{s}\right)\right) \frac{\rvx}{\|\rvx\|} 
\end{split}
\end{equation}
where $r \in \sR$ and $\rvx \in \sB_s^\textd / \{\vzero\}$.
$\lim_{s \rightarrow \infty} r \otimes_{s} \rvx  = r\, \rvx$~\citep{ungar2008analytic}.

The M\"{o}bius scalar multiplication has the following identities~\citep{ungar2008gyrovector}:
\begin{equation}
\label{eq:mobius_scalar_multiplication_properties}
\begin{split}
&1 \otimes \rva = \rva \quad \text{Identity scalar multiplication}\\
&(r_1 + r_2) \otimes \rva = r_1 \otimes \rva  \oplus  r_2 \otimes \rva  \quad \text{scalar distributive law}\\
&(r_1\,r_2) \otimes \rva = r_1 \otimes \left( r_2 \otimes \rva\right)  \quad \text{scalar associative law}\\
&\frac{|r| \otimes \rva}{\|r \otimes \rva\|} = \frac{\rva}{\|\rva\|}  \quad \text{scaling property}\\
\end{split}
\end{equation}

\paragraph{M\"obius matrix multiplication.}
\begin{equation}
\label{eq:mobius_matrix_multiplication}
\rmQ^{\otimes_{s}}(\rvx) \coloneqq 
s\, \tanh \left(\frac{\|\rmQ\rvx\|}{\|\rvx\|} \tanh ^{-1}\left(\frac{\|\rvx\|}{s}\right)\right) \frac{\rmQ\,\rvx}{\|\rmQ\, \rvx\|}, 
\end{equation}
Here, there are some properties of the M\"obius matrix multiplication~\citep{ganea2018hyperbolic}:
$\rmQ^{\otimes_s} (\rvx)  = \vzero$ if $\rmM \rvx = \vzero$.
$\rmQ \in \gM_{m, n} (\sR)$ and $\rvx \in \sB_s^\textd$.
$\rmQ^{\otimes}(\rvx)  = \rmQ^{\otimes_s}\rvx  $, then $(\rmQ^\prime\rmQ^{\otimes_s})\rvx  = \rmQ^\prime\otimes_s (\rmQ^{\otimes_s} \rvx)$.
$(r \rmQ) \otimes_s \rvx = r \otimes (\rmQ {\otimes_s} x)$ for $r \in \sR$.
$\rmQ\otimes_s \rvx = \rmQ \rvx$ for $\rmQ\transpose \rmQ = \rmI$ (rotations are preserved).  
$\lim_{s \rightarrow \infty} \rmQ\otimes_s \rvx  = \rmQ\,\rvx $.\\




\paragraph{M\"obius addition.}
For any $\rvx, \rvy \in\sB_s^{d}$, the M\"{o}bius addition (hyperbolic translation~\citep{ratcliffe2006foundations}) is defined as
\begin{equation}
\label{eq:mobius_addition}
\rvx \oplus_{s} \rvy \coloneqq \frac{\left(1 + \frac{2}{s^2} \langle\rvx, \, \rvy \rangle + \frac{1}{s^2}\|\rvy\|^2 \right)\, \rvx + \left(1 - \frac{1}{s}\|\rvx\|^2\right) \, \rvy  }{1 + \frac{2}{s^2} \langle\rvx, \, \rvy \rangle + \frac{1}{s^4} \|\rvx\|^2\|\rvy\|^2}.
\end{equation}
Moreover, we have $(-\rvx) \oplus_s \rvx = \rvx \oplus_s (-\rvx) = \vzero$ and  $(-\rvx) \oplus_s (\rvx \oplus_s \rvx) = \rvy$ (left cancellation law).
However, this operation is usually not commutative, i.e., $(\rvx \oplus_s \rvy) \oplus_s (-\rvy) \neq \rvy$~\citep{ungar2014analytic}.
$\lim_{s \rightarrow \infty} \rvx \oplus_s \rvy  = \rvx + \rvy $.\\

The table~\ref{Tab:gyrovector_addition_properties} presents the group identities of the M\"obius addition.
\begin{table*}
	\small
	\begin{tabular}{rll}
		\toprule
		Group identity && Name/reference\\
		\midrule
		$\ominus(\ominus a) = a$ && involution of inversion\\
		$\ominus a \oplus (a \oplus x) = x$&&  Left cancellation law\\
		$\gyr[a, b] c = \ominus (a \oplus b) \oplus (a \oplus (b \oplus c))$&& Gyrator identity\\
		$\ominus (a \oplus b) = \gyr[a, b](\ominus b \ominus a)$&& cf. $(ab)^{-1} = b^{-1} a^{-1}$ in a group\\
		$(\ominus a \oplus b) \oplus \gyr[\ominus a, b](\ominus b \oplus c) = \ominus a \oplus c$&& cf. $(a^{-1} b)(b^{-1}c) = a^{-1}c$ in a group\\
		$\gyr[\ominus a, \ominus b] = \gyr[a, b]$&& Even property\\
		$\gyr[b, a] = \gyr^{-1}[a, b]$&& Inversive symmetry\\
		$\phi(\gyr[a, b]c) = \gyr[\phi(a), \phi(b)]\phi(c)$&& Gyration preserving under a gyrogroup homomorphism $\phi$\\
		$L_a \circ L_b=  L_{a\oplus b}\gyr[a, b]$&& Composition law for left gyrotranslations\\
		\bottomrule
	\end{tabular}
	\caption{Group identities of Gyrovector addition~\citep{rassias2000mathematical}.}
	\label{Tab:gyrovector_addition_properties}
\end{table*}
%
%
%

\paragraph{Gyrolines.}
Let $\rvv =  (-\rvx) \oplus_s \rvy$ be a gyrovector with tail $\rvx$ and head $\rvy$.
Then, the hyperbolic line or Gyroline is defined as: 
\begin{equation}
L(t)= \rvx \oplus_s \left[ t \otimes_s ( (-\rvx) \oplus_s \rvy) \right], \quad -\infty < t < \infty.
\end{equation}
For $t \in [0, 1]$, the M\"{o}bius gyrolines are identical to the geodesics of the Poincaré disc model
of hyperbolic geometry connecting points $\rvx, \rvy \in \sB_s^\textd$~\citep{ungar2014analytic}.
$\lim_{s \rightarrow \infty} L(t)= \rvx + t\, ( \rvy -\rvx)$.\\

\paragraph{Exponential and Logarithmic map.}
We can also represent the exponential and logarithmic map using the M\"obius addition~\citep{ganea2018hyperbolic}. 
For any point $\rvx \in \sB_s^\textd$, the exponential map $\Exp_{\rvx}: T_\rvx \sB_s^\textd \rightarrow \sB_s^\textd$ 
and the logarithm map $\Log_\rvx: \sB_s^\textd \rightarrow T_\rvx \sB_s^\textd$ are given for $\rvv \neq \vzero$ and $\rvy \neq \rvx$ by:
\begin{equation}
\begin{aligned}
\Exp_{\rvx}^{s}(\rvv) =\rvx \oplus_{s}\left(s\, \tanh \left( \frac{\lambda_{\rvx}^{s}\|\rvv\|}{2s}\right) \frac{\rvv}{\|\rvv\|}\right), &&
\Log_{\rvx}^{s}(\rvy) = \frac{2s}{\lambda_{\rvx}^{s}} \tanh ^{-1}\left(\frac{\left\|-\rvx \oplus_{s} \rvy\right\|}{s}\right) \frac{-\rvx \oplus_{s} \rvy}{\left\|-\rvx \oplus_{s} \rvy\right\|}
\end{aligned}
\end{equation}

\subsection{Proofs of main theorems\label{section: proofs}}
%
We first prove two useful lemmas. 
\begin{lemma}
	For $x$ and $a \in R_+$, we have:
	\begin{equation}
	\label{eq:bound_arcsinh}
	\sinh^{-1} \frac{x}{a}  \leq \sqrt{x \frac{\pi}{2}}
	\end{equation}
	
	\begin{proof}
		\begin{equation*}
		\begin{split}
		\sinh^{-1} \frac{x}{a} &=  \int_{0}^{x} \frac{du}{\sqrt{a^2 + u^2 }}\\
		& \leq \sqrt{\int_{0}^{x} du \int_{0}^{x} \frac{du}{a^2 + u^2 }} \quad \text{(Cauchy Schwarz)}\\
		& =  \sqrt{x  \tan^{-1}\left(\frac{x}{a}\right)}, \\
		& \leq \sqrt{x \frac{\pi}{2}}, \quad 0 \leq \tanh^{-1} x \leq \pi/2, \, \text{and } x\geq 0
		\end{split}
		\end{equation*}
	\end{proof}
\end{lemma}

\begin{lemma}[Distortion of M\"obius matrix multiplication]
	Let $r \in \sR$,  $\rmL \in \sR^{n \times d}$, and $\rvw \in \sB_s^\textd/ \{\vzero\}$, then we have
	\begin{equation}
	\label{eq:distance_and_mobius_matrix_mul}
	d(\vzero, \rmL^{\otimes}\rvw) =  \frac{\|\rmL \rvw\|}{\|\rvw\|} d(\vzero, \rvw)
	\end{equation}
	\begin{proof}
		For completeness, we recall the scaling property of M\"obius scalar multiplication on hyperbolic distances~\citep{kim2013unit}.
		Let $r \in \sR$ and $\phi_{s, \rvu} = s \tanh^{-1} \left(\frac{\|\rvu\|}{s}\right)$ be the rapidity of $\rvu \in \sB_s^\textd$, 
		\begin{equation*}
		\begin{aligned}
		d_s(\vzero, r\otimes_s \rvw) &= 2s \tanh^{-1}\left(\frac{\|r\otimes_s \rvw\|}{s}\right) \\
		&=  2s\tanh^{-1}\left(\frac{1}{s}\left\|s\tanh(r\ \phi_{s, \rvw})\frac{\rvw}{\|\rvw\|}\right\|\right) && \text{(Definition of Mob\"ius scalar mult.)}\\
		& = 2s \tanh^{-1}\left(|\tanh(r\ \phi_{s, \rvw})|\right)\\
		& = 2s \tanh^{-1}\left(\tanh(|r|\ \phi_{s, \rvw})\right) && \text{(Scaling prop. Eq.~\ref{eq:mobius_scalar_multiplication_properties})}\\
		& = 2s |r| \tanh^{-1}\left(\frac{\|\rvw\|}{s}\right) \\
		& = |r|\ d_s(\vzero, \rvw)
		\end{aligned}
		\end{equation*}
		Rewritting the definition of M\"obius matrix multiplication~(Eq.\ref{eq:mobius_scalar_multiplication}) in terms of the M\"obius scalar multiplication,
		\begin{equation}
		\label{eq:mobius_matrix_mult_as_scalar_mult}
		\rmM^{\otimes_{s}}(\rvx) 
		= \left(\frac{\|\rmM \rvx\|}{\|\rvx\|} \otimes_s \rvx \right) \frac{\|\rvx\|}{\rvx}  \frac{\rmM\,\rvx}{\|\rmM\, \rvx\|} 
		\end{equation}
		Taking he norm of Eq.~\ref{eq:mobius_matrix_mult_as_scalar_mult} and 
		pluging it in Eq.~\ref{eq:distance_and_mobius_matrix_mul} gives the desired result.
	\end{proof}
\end{lemma}

\subsubsection{Hyperbolic linear layer}
%
We present the consistency of the M\"obius matrix multiplication. 
We use uniform stability to show that our approach is consistent.
An Euclidean version of this proof is presented in~\citep{perrot2015regressive} and applied to OT-mapping estimation in \citep{perrot2016mapping}.\\

We denote $\|\cdot\|_\textF$ the Frobenius norm of a matrix.
Let $\rvx_i \in \gX$, $\rvy_i \in \gY$  for $i \in [n]$, and $f: \gX \rightarrow \gY$. 
We define the loss (Eq.~\ref{eq:loss}),  the empirical risk (Eq.~\ref{eq:empirical_risk}),
the empirical regularized risk (Eq.~\ref{eq:empirical_regularized_risk}), 
the empirical risk  on truncated training set (Eq.~\ref{eq:empirical_truncated_risk}), and 
the empirical regularized risk on truncated training set (Eq.~\ref{eq:empirical_regularized_truncated_risk}).
\begin{align}
\ell(f,\, (\rvx, \rvy)) &= d_s(f(\rvx),\, \rvy), \label{eq:loss} && \text{(Hyperbolic distance)}\\
\hat{R}(f) &= \frac{1}{n} \sum_{j=1}^{n} \ell(f,\, (\rvx_j, \rvy_j)), \label{eq:empirical_risk}\\
\hat{R}^{\backslash i}(f) &= \frac{1}{n} \sum_{j \neq i}^{n} \ell(f,\, (\rvx_j, \rvy_j)), \label{eq:empirical_truncated_risk}\\
\hat{R}_r(f) &= \frac{1}{n} \sum_{j = 1}^{n} \ell(f,\, (\rvx_j, \rvy_j)) + \omega\, \Omega(f), \label{eq:empirical_regularized_risk}\\
\hat{R}^{\backslash i}_r(f) &= \frac{1}{n} \sum_{j \neq i}^{n} \ell(f,\, (\rvx_j, \rvy_j)) + \omega\, \Omega(f)  \label{eq:empirical_regularized_truncated_risk},
\end{align}
where $\Omega(\cdot)$ is the regularization term, and $\omega \geq 0$ is the regularization parameter.

The quality loss of the hyperbolic mapping estimation is:
\begin{equation}
\label{eq:quality_loss}
T \leftarrow \argmin_{T \in \gT} g(T) \coloneqq \frac{1}{n} \sum_{i=1}^{n} \ell\left(T,\, (\rvx_i, \rvy_i)\right) + \omega\, \Omega(T),
\end{equation}
where $\rvy_i = B_{\rmM}^\hyp(\rvx_i)$ for $i \in [n]$.

We use the set of hyperbolic linear transformations induced by a real matrix $\rmL \in \sR^{\text{d} \times \text{p}}$ in this analysis:
\begin{equation}
\gT = \left\{ T: \exists \rmL \in \sR^{\text{d} \times \text{p}}, \, \forall \rvx \in \gX, \, T(\rvx) = \left(\rmL^{\otimes_s} \rvx \right)^\top  \right\}.
\end{equation}

%
In our setting, let $\rvx \in \sB_s^\text{d}$ and $\rvv  \in \sB_s^\text{d}$ be in the Poincaré ball of $\text{d}$ dimensions and radius $s$.
We assume $\|\rvx\|\leq C_x  < s$ and $\|\rvy\|\leq C_y \leq s$.
Then, using the definition of hyperbolic distances to zero Eq.~\ref{eq:hyperbolic_distace_to_zero}, we have $d_s(\vzero,\, \rvx) \leq d_s(\vzero,\, C_x) = K_x$ and $d_s(\vzero,\, \rvy) \leq d_s(\vzero,\, C_y) = K_y$,  as the $\tanh^{-1}(\cdot)$ is monotonically increasing.\\


\begin{lemma}\label{lemma:bound_regularizer}
	Let $\rmL$ be an optimal solution of Problem~\ref{eq:quality_loss}, we have: 
	\begin{equation}
	\|\rmL\|_F \leq \frac{K_y}{\omega}
	\end{equation}
	\begin{proof}
		\begin{equation*}
		\begin{aligned}
		g(\rmL) &\leq g(\vzero)\\
		\hat{R}(\rmL) + \omega\|\rmL\|_F &\leq \hat{R}(\vzero) + \omega\|\vzero\|_F && \text{(optimality of $\rmL$)}\\
		\omega\|\rmL\|_F &\leq \frac{1}{n} \sum_{j=1}^{n}d_s(\vzero, \rvv_j) && \text{(problem~\ref{eq:quality_loss}  is always positive)}\\
		\|\rmL\|_F &\leq \frac{K_y}{\omega} && (d_s(\vzero, \, \rvv) \leq K_y)\\
		\end{aligned}
		\end{equation*}
	\end{proof}
	
\end{lemma}

\begin{lemma}\label{lemma:bound_loss}
	The loss $\ell(\rmL, \, (\rvx, \rvv))$ is bounded by $M = K_y \left( \frac{K_x}{\omega} + 1\right) $
	
	\begin{proof}
		\begin{equation*}
		\begin{aligned}
		\ell(\rmL, \, (\rvx, \rvv)) &= d_s(\rvv, \, \rmL^\otimes\rvx) &&\\
		&\leq d_s(\vzero, \, \rmL^\otimes\rvx) + d_s(\vzero, \, \rvv) && \text{(Triangle ineq.)}\\
		&= \frac{\|\rmL\rvx\|}{\|\rvx\|}d_s(\vzero, \, \rvx) + d_s(\vzero, \, \rvv) && \text{(Eq.~\ref{eq:distance_and_mobius_matrix_mul})}\\
		&\leq \|\rmL\|_F\, d_s(\vzero, \, \rvx) + d_s(\vzero, \, \rvv) && \text{(Prop. of norms)}\\
		&\leq \|\rmL\|_F\, K_x + K_y && (d_s(\vzero, \, \rvx) \leq K_x,\,  d_s(\vzero, \, \rvv)\leq K_y)\\
		&\leq K_y \left( \frac{K_x}{\omega} + 1\right)  && \text{(Lemma~\ref{lemma:bound_regularizer})}\\
		\end{aligned}
		\end{equation*}
	\end{proof}
\end{lemma}

\begin{lemma}
	\label{lemma:upper_bound_lorentz_factor_mobius_mat_mult}
	Upper bound of the Lorentz gamma factor of $\rmP^\otimes \rvx$ for $\rvx \in \sB_s^\textd$ where 
	$d_s(\vzero, \, \rvx) \leq K_x$:  
	\begin{equation}
	\label{eq:bound_gamma_of_hyperbolic_matrix_multiplication}
	\gamma_{\rmP^\otimes \rvx}  \leq \exp\left(\frac{\|\rmP\|_F^2}{s^2} \frac{K_x^2}{8} \right).
	\end{equation}
	
	\begin{proof}
		\begin{equation*}
		\begin{aligned}
		\gamma_{\rmP^\otimes \rvx} = \frac{1}{\sqrt{1 - \frac{\|\rmP^\otimes \rvx\|^2}{s^2}}} &\leq \frac{1}{\sqrt{1 - \tanh^2\left(\frac{\|\rmP\|_F}{s} \frac{K_x}{2} \right) }} && \text{(Eq.~\ref{eq:bound_norm_hyperbolic_matrix_multiplication})}\\
		& = \cosh\left(\frac{\|\rmP\|_F}{s} \frac{K_x}{2} \right) \\
		& \leq \exp\left(\frac{\|\rmP\|_F^2}{s^2} \frac{K_x^2}{8} \right)  && (\cosh x \leq \exp(x^2 / 2))
		\end{aligned}
		\end{equation*}
	\end{proof}
\end{lemma}

\begin{lemma}
	Let $\|\rvx\| \leq C_x$,  $\rmP$ and $\rmQ$ in $\sR^{n\times n}$.
	If the matrices $\rmP$ and $\rmQ$ satisfy $\frac{1}{\|\rmP\rvx\|}, \frac{1}{\|\rmQ\rvx\|}  \leq L$, then
	\begin{equation}
	\label{eq:bound_diff_hyperbolic_mult}
	\|\rmP^\otimes\rvx  - \rmQ^\otimes\rvx \|  \leq 2 s\, L C_x\|\rmP - \rmQ\|_F. 
	\end{equation}
	
	\begin{proof}
		Let us define 
		\begin{eqnarray*}
			k_{1}(\rvx) &= \left(\frac{\|\rmP\rvx\|}{\|\rvx\|} \otimes \|\rvx\| \right)\frac{1}{\|\rmP \rvx\|} \leq s\frac{1}{\|\rmP \rvx\|}, \\
			k_{2}(\rvx) &= \left(\frac{\|\rmQ\rvx\|}{\|\rvx\|} \otimes \|\rvx\| \right)\frac{1}{\|\rmQ \rvx\|} \leq s\frac{1}{\|\rmQ \rvx\|}.
		\end{eqnarray*}
		
		We assume w.l.o.g $k_2(\rvx) > k_1(\rvx)$
		\begin{equation*}
		\begin{aligned}
		\|\rmP^\otimes\rvx & - \rmQ^\otimes\rvx \| \\
		& = \left\| k_{1}(\rvx)(\rmP\rvx - \rmQ\rvx) -  (k_{2}(\rvx) - k_{1}(\rvx))\rmQ\rvx \right\| && \text{(Rewritting)}\\
		& \leq k_{1}(\rvx) \left\|\rmP\rvx - \rmQ\rvx\| + (k_{2}(\rvx) - k_{1}(\rvx)) \| \rmQ\rvx \right\| && \text{(Triangle ineq.)}\\
		& \leq s \left| \frac{1}{\|\rmP\rvx\|} \|\rmP\rvx - \rmQ\rvx\| + \left(\frac{1}{\|\rmQ\rvx\|}  - \frac{1}{\|\rmP\rvx\|} \right) \| \rmQ\rvx \|\right| && \text{(Subadditivity of $|\cdot|$)}\\
		%
		& \leq \frac{s}{\|\rmP\rvx\|} \left| \|\rmP\rvx - \rmQ\rvx\| \right|+ \left|\|\rmP\rvx\| - \|\rmQ\rvx\|   \right|\\
		& \leq \frac{2 s}{\|\rmP\rvx\|} \|\rmP\rvx - \rmQ\rvx\| && \text{(Reverse trinagle ineq.)}\\
		& \leq 2 s\, L \|\rmP\rvx - \rmQ\rvx\| && (\frac{1}{\|\rmP \rvx\|} \leq L)\\
		& \leq 2 s\, L C_x\|\rmP - \rmQ\|_F && \text{(Prop. of norms and $\|\rvx\| \leq C_x$)}\\
		\end{aligned}
		\end{equation*}
	\end{proof}
\end{lemma}

\begin{lemma}[Upper bound of the norm the M\"obius matrix multiplication]
	Let $\rvx \in \sB_s^\textd$ with $\|\rvx\| \leq C_x$ and $d_s(\vzero, \, \rvx) \leq K_x$.
	Then,  
	\begin{equation}
	\label{eq:bound_norm_hyperbolic_matrix_multiplication}
	\|\rmP^\otimes \rvx\|  \leq s \tanh\left(\frac{\|\rmP\|_F}{s} \frac{K_x}{2} \right).
	\end{equation}
	
	\begin{proof}
		\begin{equation*}
		\begin{aligned}
		\|\rmP^\otimes \rvx\| &= \frac{\|\rmP \rvx\|}{\|\rvx\|} \otimes \|\rvx\|  && \text{(Rewritting)}\\
		&= s \tanh\left(\frac{\|\rmP\rvx\| }{\|\rvx\|} \tanh^{-1}\left(\frac{\|\rvx\|}{s} \right) \right) \\
		& \leq s \tanh\left(\|\rmP\rvx\| \frac{K_x}{2 s \|\rvx\|} \right) && (d(\vzero, \, \rvx) \leq K_x)\\
		&\leq s \tanh\left(\frac{\|\rmP\|_F}{s} \frac{K_x}{2} \right). && \text{(Prop. of norms)}\\
		%
		\end{aligned}
		\end{equation*}
	\end{proof}
\end{lemma}

\begin{lemma}[\label{lemma:bound_loss_hyperbolic_mat_mul} $\ell_\infty$ bound of the hyperbolic loss]
	Let $\rvx$, $\rvv$ $\in \sB_s^\textd \backslash \{\vzero\}$ with $\|\rvx\| \leq C_x$, and $\rmP$, $\rmQ$ in $\sR^{n\times d}$, 
	where $\rmP$ and $\rmQ$ satisfy $\frac{1}{\|\rmP\rvx\|}, \frac{1}{\|\rmQ\rvx\|}  \leq L$, then
	
	\begin{equation}
	\left| \ell(\rmP, \,  (\rvx, \rvv)) - \ell(\rmQ, \,(\rvx, \rvv))\right|
	\leq \sqrt{2\pi s^3 L C_x} \, \left(\|\rmP - \rmQ\|_F + 1\right) 
	\end{equation}

	\begin{proof}
		Note that 
		\begin{equation}
		\left| \ell(\rmP, \,  (\rvx, \rvv)) - \ell(\rmQ, \,(\rvx, \rvv))\right| \leq d(\rmP^\otimes\rvx, \rmQ^\otimes\rvx) \quad \text{(Reverse triangle ineq.~\citep{demirel2008hyperbolic})}
		\end{equation}
		%
		%
		\begin{equation*}
		\begin{aligned}
		d(\rmP^\otimes\rvx, \, \rmQ^\otimes \rvx)  &= 2s\, \sinh^{-1}\left(\gamma_{\rmP^\otimes \rvx}\gamma_{\rmQ^\otimes \rvx}\frac{\|\rmP^\otimes \rvx - \rmQ^\otimes \rvx \|}{s}\right) && \text{(Eq.~\ref{eq:hyperbolic_distances})}\\
		& \leq 2s\, \sinh^{-1}\left(\exp\left(\frac{\|\rmP\|_F^2 + \|\rmQ\|_F^2}{s^2} \frac{K_x^2}{8} \right) \frac{\|\rmP^\otimes \rvx - \rmQ^\otimes \rvx \|}{s}\right)&& \text{(Eq.~\ref{eq:bound_gamma_of_hyperbolic_matrix_multiplication})}\\
		& \leq 2s\, \sqrt{\frac{\pi}{2} \|\rmP^\otimes \rvx - \rmQ^\otimes \rvx \| } && \text{(Eq.~\ref{eq:bound_arcsinh})} \\
		& \leq 2s\,  \sqrt{2\pi s L C_x \|\rmP - \rmQ\|_F} && \text{(Eq.~\ref{eq:bound_diff_hyperbolic_mult})} \\
		& \leq \sqrt{2\pi s^3 L C_x} \, \left(\|\rmP - \rmQ\|_F + 1\right) && 
		(\sqrt{a} \leq (a + 1)/2)
		\end{aligned}
		\end{equation*}
	\end{proof}
\end{lemma}

\begin{lemma}[Majorizer of the hyperbolic distance of linear combination of matrices in M\"obius matrix multiplication]
	\label{lemma:majorizer}
	Let $\rvx$ and $\rvv$ in $\sB_s^\textd$ with $d_s(\vzero, \, \rvv) \leq K_y$.  For $t \in [0, 1]$ we have:
	\begin{equation}
	d(\rvv,\, \left[t \rmP + (1 - t)\rmQ\right]^{\otimes_s} \rvx) \leq  t\, d(\rvv,\, \rmP^{\otimes_s} \rvx) + (1 - t)\, d(\rvv,\, \rmQ^{\otimes_s} \rvx) + 2K_y
	\end{equation}
	
	\begin{proof}
		\begin{equation*}
		\begin{aligned}
		d(\rvv,\, & \left[t \rmP + (1 - t)\rmQ\right]^{\otimes_s} \rvx)  \\
		&\leq \left|d(\vzero,\, \left[t \rmP + (1 - t)\rmQ\right]^\otimes \rvx) + d(\vzero,\, \rvv)\right|, && \text{(Triangle ineq.)}\\
		&= \left|\frac{\|\left[t \rmP + (1 - t)\rmQ\right]\rvx\|}{\|\rvx\|} d(\vzero,\, \rvx) + d(\vzero,\, \rvv)\right|, && \text{(Eq.~\ref{eq:distance_and_mobius_matrix_mul} )}\\
		&\leq \left|t\, d(\vzero,\, \rmP^{\otimes_s} \rvx) + (1 - t)\,d(\vzero,\, \rmQ^{\otimes_s} \rvx) +  d(\vzero,\, \rvv) \right| && (\text{Convexity of $\|\cdot \|$}) \\
		&\leq \left| t \left[d(\vzero,\, \rmP^{\otimes_s} \rvx) - d(\vzero,\, \rvv)\right] + (1 - t)\left[d(\vzero,\, \rmQ^{\otimes_s} \rvx) - d(\vzero,\, \rvv)\right]\right| +  \left|2d(\vzero,\, \rvv)\right|  && \text{(Subadditivity of $|\cdot|$)}\\
		&\leq \left| t\, d(\rvv,\, \rmP^{\otimes_s} \rvx) + (1 - t)\, d(\rvv,\, \rmQ^{\otimes_s} \rvx) \right| +  \left|2d(\vzero,\, \rvv)\right|  && \text{(Reverse triangle ineq.)}\\
		&\leq  t\, d(\rvv,\, \rmQ^{\otimes_s} \rvx) + (1 - t)\, d(\rvv,\, \rmQ^{\otimes_s} \rvx) +  2 K_y 
		\end{aligned}
		\end{equation*}
	\end{proof}
\end{lemma}

\begin{lemma}\label{lemma:bound_regularized_diff}
	Let $T(\rvx) = \rmL^\otimes \rvx$.	
	We denote $\rmL$ a minimizer of $\hat{R}_r$ and for $i \in [n]$, let $\rmL^i$ denote a minimizer of $\hat{R}^{\backslash i}_r$.
	Then, 
	\begin{equation}
	\|\rmL\|_F^2 - \|\rmL^i + t \,\Delta\rmL\|_F^2 + \|\rmL^i\|_F^2 - \|\rmL^i - t\, \Delta\rmL\|_F^2 \leq \frac{\sqrt{8\pi s^3 L C_x}}{\omega n} \, \left(t \|\Delta \rmL\|_F + 1\right) + 
	\frac{4 K_y}{\omega n}
	\end{equation}
	
	\begin{proof}
		Let $\hat{R}_r$ and $\hat{R}^{\backslash i}_r$ be the functions to optimize, $\rmL$ and $\rmL^i$ their respective minimizers and $\omega$ the regularization parameter used in our algorithm. 
		Let $\Delta \rmL = \rmL - \rmL^i$, then applying Lemma~\ref{lemma:majorizer} for $t \in [0, 1]$:
		\begin{align}
		\hat{R}^{\backslash i}_{\text{emp}}(\rmL - t\Delta \rmL)  & \leq   t\, \hat{R}_{\text{emp}}^{\backslash i}(\rmL^i) + (1 - t) \hat{R}_{\text{emp}}^{\backslash i}(\rmL)  +  \frac{2 K_y}{n}  \label{eq:solution_L}\\
		\hat{R}^{\backslash i}_{\text{emp}}(\rmL^i + t\Delta\rmL)  & \leq   t\, \hat{R}^{\backslash i}_{\text{emp}}(\rmL) + (1 - t) \hat{R}^{\backslash i}_{\text{emp}} (\rmL^i)  +  \frac{2 K_y}{n}  \label{eq:solution_Li}
		\end{align}
		Summing both inequalities,
		\begin{equation}
		\label{eq:loo_risk}
		\hat{R}^{\backslash i}_{\text{emp}}(\rmL - t\Delta\rmL) + \hat{R}^{\backslash i}_{\text{emp}}(\rmL^i + t\Delta\rmL) -  \hat{R}^{\backslash i}_{\text{emp}}(\rmL)  -  \hat{R}^{\backslash i}_{\text{emp}}(\rmL^i) -  \frac{4 K_y}{n}  \leq 0
		\end{equation}
		
		Let $\rmL$ and $\rmL^i$ be minimizer of Eq.~\ref{eq:solution_L} and Eq.~\ref{eq:solution_Li}, then 
		\begin{equation}
		\label{eq:risk_assumptions}
		\begin{split}
		\hat{R}_r(\rmL) - \hat{R}_r(\rmL - t \,\Delta\rmL) \leq 0,\\
		\hat{R}^{\backslash i}_r (\rmL^i)  - \hat{R}^{\backslash i}_r (\rmL^i + t\, \Delta\rmL)  \leq 0
		\end{split}
		\end{equation}

		Adding inequalities in Eq.~\ref{eq:risk_assumptions} and Eq.~\ref{eq:loo_risk},
		\begin{equation*}
		\begin{aligned}
		\hat{R}_{\text{emp}}(\rmL) &- \hat{R}_{\text{emp}}(\rmL -  t \,\Delta\rmL) + \hat{R}^{\backslash i}_{\text{emp}}(\rmL - t\, \Delta\rmL) -
		\hat{R}^{\backslash i}_{\text{emp}}(\rmL)\\
		& \omega\|\rmL\|_F^2 - \omega\|\rmL^i + t \,\Delta\rmL\|_F^2 + \omega\|\rmL^i\|_F^2 - \omega\|\rmL^i - t\, \Delta\rmL\|_F^2 - \frac{4 K_y}{n}  \leq 0
		\end{aligned}
		\end{equation*}
		Then, 
		\begin{equation*}
		\|\rmL\|_F^2 - \|\rmL^i + t \,\Delta\rmL\|_F^2 + \|\rmL^i\|_F^2 - \|\rmL - t\, \Delta\rmL\|_F^2 \leq \frac{B}{ \omega} +  \frac{4 K_y}{\omega n}
		\end{equation*}
		
		\begin{equation*}
		\begin{aligned}
		B  \leq & \left| 
		\hat{R}_{\text{emp}}^{\backslash i}(\rmL) -
		\hat{R}_{\text{emp}}^{\backslash i}(\rmL - t\, \Delta\rmL) +
		\hat{R}_{\text{emp}}(\rmL - t \,\Delta\rmL) -
		\hat{R}_{\text{emp}}(\rmL) 
		\right|\\
		= & \left| 
		\frac{1}{n} \sum_{(\rvx, \rvv)^i \in S^i} \ell(\rmL, \,  (\rvx, \rvv)^i)  - 
		\frac{1}{n} \sum_{(\rvx, \rvv)^i \in S^i} \ell(\rmL + t\, \Delta\rmL, \,  (\rvx, \rvv)^i) \right.\\
		& +
		\left. \frac{1}{n} \sum_{(\rvx, \rvv) \in S} \ell(\rmL + t\, \Delta\rmL, \,  (\rvx, \rvv) ) -
		\frac{1}{n} \sum_{(\rvx, \rvv) \in S} \ell(\rmL,\,  (\rvx, \rvv)) 
		\right|\\
		\leq & \frac{1}{n} 
		\left| 
		\ell(\rmL,\,  (\rvx_i, \rvv_i))  - 
		\ell(\rmL + t\, \Delta\rmL, \,  (\rvx_i, \rvv_i))  +
		\ell(\rmL,\,  (\rvx_i, \rvv_i)^i)  -
		\ell(\rmL + t\, \Delta\rmL, \,  (\rvx_i, \rvv_i)^i) 
		\right| 
		\\
		\leq & \frac{1}{n} 
		\left| 
		\ell(\rmL,\,  (\rvx_i, \rvv_i))  - 
		\ell(\rmL + t\, \Delta\rmL, \,  (\rvx_i, \rvv_i)) 
		\right| 
		+
		\frac{1}{n} \left| 
		\ell(\rmL,\,  (\rvx_i, \rvv_i)^i) - 
		\ell(\rmL + t\, \Delta\rmL, \,  (\rvx_i, \rvv_i)^i) 
		\right|\\
		\leq & \frac{\sqrt{8\pi s^3 L C_x}}{n} \, \left(t \|\Delta \rmL\|_F + 1\right)
		\end{aligned}
		\end{equation*}
		
	\end{proof}
\end{lemma}
where the second inequality comes from the fact that the sums differ by the $i$th-element.
The last inequality uses lemma~\ref{lemma:bound_loss_hyperbolic_mat_mul}.

\begin{definition}[Uniform stability~\citep{bousquet2002stability}]
	An algorithm $A$ has uniform stability $\beta$ with respect to the loss function $\ell$ if the following holds
	\begin{equation*}
	\forall S \in D, \forall i \in [n], \, \|\ell(A_S, \cdot ) - \ell(A_{S^{\backslash i}}, \cdot) \|_\infty \leq \beta.
	\end{equation*}
\end{definition}

\begin{lemma}\label{lemma:uniform_stability}
	Our algorithm has uniform stability 
	$\beta = \frac{1}{2 \omega n}\left(\frac{N_x^2}{4\omega  n} + 2N_x + 8 K_y + 1\right)$, where $N_x =  \sqrt{2\pi s^3 LC_x}$.

	\begin{proof}
		Set $t=\frac{1}{2}$ on the left hand side of lemma~\ref{lemma:bound_regularized_diff}, 
		\begin{equation*}
		\|\rmL\|_F^2 - \|\rmL^i + \frac{1}{2}\Delta\rmL\|_F^2 + \|\rmL^i\|_F^2 - \|\rmL - \frac{1}{2}\Delta\rmL\|_F^2 = \frac{1}{2}\|\Delta\rmL\|_F^2
		\end{equation*}
		and thus:
		\begin{equation*}
		\begin{aligned}
		\frac{1}{2}\|\Delta\rmL\|_F^2 & \leq \frac{\sqrt{2\pi s^3 L C_x}}{\omega n} \, \left(\|\Delta \rmL\|_F + 1\right) + \frac{4 K_y}{\omega n}\\
		%
		%
		%
		\|\Delta\rmL\|_F & \leq \frac{N_x}{2 \omega n} + \frac{1}{\omega n}\sqrt{\frac{N_x^2}{4\omega  n} + N_x + 8 K_y},   && \text{where } N_x = \sqrt{2\pi\, s^3 LC_x}. \\
		& \leq \frac{1}{2 \omega n}\left(\frac{N_x^2}{4\omega  n} + 2N_x + 8 K_y + 1\right)&& (\sqrt{a} \leq (a + 1) / 2 )
		\end{aligned}
		\end{equation*}
	\end{proof}
\end{lemma}

\begin{theorem}[\citep{bousquet2002stability}\label{thm:stability}]
	Let $A$ be an algorithm with uniform stability $\beta$ with respect to a loss function $l$ such that 
	$0 \leq l(A_S, z) \leq M$, for all $z \in \gZ$ and all sets $S$.
	Then, for any $n \geq 1$, and any $\delta \in (0, 1)$, the following bounds hold (separately) with 
	probability $1 - \delta$ over the random draw of the sample $S$
	\begin{equation*}
	\begin{split}
	R & \leq \hat{R}_{\text{emp}} + 2 \beta + (4n\beta + M ) \sqrt{\frac{\ln 1/\delta}{2n}},\\
	\end{split}
	\end{equation*}
\end{theorem}

\begin{theorem}
	Let $\|\rvy\| \leq C_y$ and $\|\rvx\| \leq C_x$, for any $\rvx$ and $\rvy$ in $\sB_s^\textd\backslash \{\vzero\}$
	with a probability of $1 - \delta$  for any matrix $\rmL$ optimal solution of problem~\ref{eq:quality_loss} such that $\|\rmL \rvx\|^{-1} \leq L$ and $\|\rmL \rvy\|^{-1} \leq L$, we have:
	\begin{equation}
	\label{eq:bounds_risk}
	\begin{split}
	R(\rmL) \leq & \hat{R}_{\text{emp}}(\rmL) + \frac{1}{\omega n}\left(\frac{N_x^2}{4\omega  n} + 2N_x + 8 K_y + 1\right) \\
	& + 
	\left(\frac{2}{ \omega }\left(\frac{N_x^2}{4\omega  n} + 2N_x + 1\right) + K_y\left(\frac{K_x + 16}{\omega}  + 1 \right)
	\right) \sqrt{\frac{\ln(1/\delta)}{2 n}},
	\end{split}
	\end{equation}
	where $N_x = 2s\, \sqrt{2\pi s LC_x}$.

	\begin{proof}
		As our algorithm is uniformly stable and our loss is bounded, hence we can apply Theorem~\ref{thm:stability},   
		with $M=K_y\left(\frac{K_x}{\omega}  + 1 \right)$ (lemma~\ref{lemma:bound_loss}), and 
		$\beta = \frac{1}{2 \omega n}\left(\frac{N_x^2}{4\omega  n} + 2N_x + 8 K_y + 1\right)$, where $N_x =  \sqrt{2\pi s^3 LC_x}$ (lemma~\ref{lemma:uniform_stability}).
	\end{proof}
\end{theorem}

\subsubsection{Hyperbolic linear layer for mapping estimation}

\begin{theorem}
	Let $f^*$ be the true transport map.
	Let $B_{\rmM_0}(\rvx)$ be the true barycentric mapping associated with the probabilistic coupling $\rmM_0$.
	Let $B_{\hat{\rmM}}(\rvx)$ be the empirical barycentric mapping of $\rmX^s$ using the probabilistic 
	coupling $\hat{\rmM}$ learned between $\rmX^s$ and $\rmX^t$.
	
	\begin{equation}
	\begin{split}
	\E_{\rvx^s \sim \gX^\texts}\left[d\left(f(\rvx^\texts), \, f^*(\rvx^\texts)\right)\right] \leq & \sum_{\rvx^\texts \in \gX^\texts} d\left(f(\rvx^\texts), \, B_{\hat{\rmM}}(\rvx^\texts)\right)  + \sum_{\rvx^\texts \in \gX^\texts} d\left(B_{\hat{\rmM}}(\rvx^s), \, B_{\rmM_0}(\rvx^\texts)\right)  \\
	& + O\left(\frac{1}{\sqrt{n^\texts}}\right)
	+  \E_{\rvx^\texts \sim \gX^\texts}\left[d\left(f^*(\rvx^\texts),\, B_{\rmM_0}(\rvx^\texts)\right)\right]
	\end{split}
	\end{equation}

	\begin{proof}
		\begin{equation*}
		\begin{aligned}
		\E_{\rvx^\texts \sim \gX^\texts} & \left[d\left(f(\rvx^\texts), \, f^*(\rvx^\texts)\right)\right] \\
		&\leq \E_{\rvx^\texts \sim \gX^\texts}\left[d\left(f(\rvx^s), \, B_{\rmM_0}(\rvx^\texts)\right)\right]
		+\E_{\rvx^\texts \sim \gX^\texts}\left[d\left(f^*(\rvx^\texts),\, B_{\rmM_0}(\rvx)\right)\right]\quad \text{(Triangle ineq.)}\\
		& \leq \sum_{\rvx^\texts \in \gX^\texts}  d(f(\rvx^\texts), \, B_{\rmM_0}(\rvx^\texts)) + O\left(\frac{1}{\sqrt{n^\texts}}\right)+ \E_{\rvx^\texts \sim \gX^\texts}\left[d(f^*(\rvx^\texts),\, B_{\rmM_0}(\rvx^\texts))\right]\quad \text{(Eq.~\ref{eq:bounds_risk})}\\ 
		& \leq \sum_{\rvx^\texts \in \gX^\texts} d\left(f(\rvx^\texts), \, B_{\hat{\rmM}}(\rvx^s)\right)  + \sum_{\rvx^\texts \in \gX^\texts} d\left(B_{\hat{\rmM}}(\rvx^\texts), \, B_{\rmM_0}(\rvx^\texts)\right)   + O\left(\frac{1}{\sqrt{n^\texts}}\right)\\
		&\quad +  \E_{\rvx^\texts \sim \gX^\texts}\left[d\left(T^*(\rvx^\texts),\, B_{\rmM_0}(\rvx^\texts)\right)\right]\quad \text{(Triangle ineq.)}
		\end{aligned}
		\end{equation*}
	\end{proof}
\end{theorem}

\paragraph{Link with Wrapped Gaussian Distributions.}

\begin{definition}
	Let $\nu$ be a measure on $X$ and $f: X\rightarrow Y$ be a measurable map.
    We define the push-forward as $f^{\sharp}\nu(A) \coloneqq \nu(f^{-1}1(A))$ for any measurable set $A$ in $Y$~\citep{mallasto2018wrapped}.
\end{definition}

\begin{definition}[Wrapped Gaussian~\citep{nagano2019wrapped, mardia2009directional,said2019warped}\label{def:wrapped_gaussian}] 
	Define $\rvx \sim \left(\Exp_{\mu}\right)_{\sharp} \gN(0, \Sigma)$ to be a wrapped Gaussian random variable with
	bias $\mu \in \sB_s^\textd$ and covariance matrix $\Sigma \in \sR^{n\times n}$.
	To build a wrapped Gaussian, we draw samples at random from a zero-mean Gaussian distribution 
	with covariance $\Sigma$, $\rvz \sim \gN(0, \Sigma)$.
	We project these samples onto the manifold at zero, $\bar{\rvx}_i = \Exp_{\vzero}(\rvz_i)$. 
	Finally, we add the bias term $\mu$ using the M\"obius addition,  $\rvx_i = \Exp_{\mu}(\rvz_i)$.
\end{definition}

\begin{theorem}[Optimal transport: Wrapped linear case]
	Let $\rvx \sim \left(\Exp_{\mu_1}^s\right)^{\sharp} \gN(0, \Sigma_1)$ and 
	$\rvy \sim \left(\Exp_{\mu_2}^s\right)^{\sharp} \gN(0, \Sigma_2)$ be two hyperbolic random variables, distributed under wrapped Gaussian 
	with parameters  $\mu_i \in \sB_s^\textd$ for $i = 1,2$ and $\Sigma_i \in \sR^{n \times n}$ for $i = 1, 2$. 
	Then,  
	\begin{equation}
	\begin{bmatrix}
	\rvx\\
	\rvy
	\end{bmatrix} = 
	\begin{bmatrix}
	\rvx\\
	\mu_2 \oplus_s \rmT^\otimes (-\mu_1 \oplus_s \rvx)
	\end{bmatrix} \sim \left(\Exp_{[\mu_1, \mu_2]}^s\right)^{\sharp}
	\gN \left(
	\vzero, 
	\begin{bmatrix}
	\Sigma_1 &  \rmT\, \Sigma_1\\
	\Sigma_1\, \rmT & \Sigma_2\\
	\end{bmatrix}
	\right),
	\end{equation}
	where $\rmT \in \Sym^+(n)$ is the solution of the Riccati equation $\rmT\,\Sigma_1\, \rmT = \Sigma_2$~\citep{bhatia2019bures,flamary2019concentration}.
	\begin{proof} 
		By the definition~\ref{def:wrapped_gaussian}, let $\rvx = \mu_1 \oplus_{c} \bar{\rvx}$ and $\rvy = \mu_2 \oplus_{c} \bar{\rvy}$ be Gyrovectors in $\sB_s^\textd$, 
		where $\bar{\rvx}$ and $\bar{\rvy}$ are zero-mean wrapped-Gaussian vectors, 
		$\bar{\rvx} \sim \left(\Exp_{\vzero}^c\right)^{\sharp}\gN(\vzero, \Sigma_1)$ and  $\bar{\rvy} \sim \left(\Exp_{\vzero}^c\right)_{\#}\gN(\vzero, \Sigma_2)$ respectively.
		Thus, $\bar{\rvx} = (-\mu_1) \oplus_{c} \rvx $ by the left-cancellation law.
		Then, we solve the transport problem in the tangent space at $\vzero$, $T_{\vzero}\sB_s^\textd$ for 
		$\rvu=\Log_{\vzero}^s(\bar{\rvx})$ and $\rvv=\Log_{\vzero}^s(\bar{\rvy})$,  which produces  $\rvv = \rmT\, \rvu$.
		Finally, we use the exponential map at zero on both sides and add the bias term $\mu_2$ on the left, obtaining $\rvy = \mu_2 \oplus_{c} \rmT^{\otimes_{s}}\, (-\mu_1 \oplus_s \rvx)$. 
	\end{proof}
\end{theorem}

\paragraph{Relationship between the Gyrobarycenter and the Barycenter.}
As we stated it in the main manuscript, we defined the matrix form of the gyrobarycenter as: 
\begin{equation}
\begin{aligned}
B_{\rmM}^\hyp (\rmX^\texts) = \frac{1}{2} \otimes_s  \diag\left(\rmM\, \rvg \right)^{-1}\,\rmM\, \rmG \, \rmX^\textt,
\end{aligned}
\end{equation}
where $\rvg = (\gamma_{\rmX}^s)^2 - \frac{1}{2}$, $\rmG = \diag((\gamma_{\rmX}^s)^2)$, 
and $\gamma_{\rmX}^s$ denotes the Lorentz gamma factor applied sample-wise.
We know that $\lim\limits_{s \rightarrow \infty }\gamma_{\rvx}^s  = 2$, then $\lim\limits_{s\rightarrow \infty} \rmG  = 4\, \rmI$, where $\rmI$ is the identity matrix.
Similarly, $\lim\limits_{s \rightarrow \infty} \rvg = \frac{7}{2} \, \vone$.
Finally, we know that $\lim\limits_{s \rightarrow \infty } r \otimes_s \rvx = r\, \rvx$ (see Section~\ref{sec:properties}).
Hence, 
\begin{equation}
\lim\limits_{s \rightarrow \infty} B_{\rmM}^{\hyp} (\rmX^\texts) = \frac{4}{7}\, B_{\rmM}^{\euc} (\rmX^\texts)
\end{equation}